\documentclass{article}

\PassOptionsToPackage{numbers,sort}{natbib}
\usepackage[preprint]{neurips_2026}

\usepackage[utf8]{inputenc} 
\usepackage[T1]{fontenc}    
\usepackage{hyperref}       
\usepackage{url}            
\usepackage{booktabs}       
\usepackage{amsfonts}       
\usepackage{nicefrac}       
\usepackage{microtype}      
\usepackage{xcolor}         
\usepackage{graphicx}
\usepackage{multirow}
\usepackage{amsmath}
\usepackage{amssymb}
\usepackage{amsfonts}
\usepackage{makecell}
\usepackage{graphicx}
\usepackage{subcaption}
\usepackage{wrapfig}
\usepackage{bbm}

\usepackage[table]{xcolor}   
\usepackage{colortbl}        

\usepackage{listings}        
\usepackage{tcolorbox}       
\usepackage{xcolor}          
\tcbuselibrary{breakable,skins}

\definecolor{CreateBlue}{HTML}{5579E1}
\definecolor{UpdateGreen}{HTML}{5F8E5F}
\definecolor{DeletePurple}{HTML}{9F56CF}

\newtcolorbox{createprompt}{
    colback=CreateBlue!10!white, 
    colframe=CreateBlue,         
    boxrule=0.6pt,
    arc=2mm,
    left=2mm,
    right=2mm,
    top=1mm,
    bottom=1mm,
    fonttitle=\bfseries,
    title=Create Prompt,
}

\newtcolorbox{updateprompt}{
    colback=UpdateGreen!10!white,
    colframe=UpdateGreen,
    boxrule=0.6pt,
    arc=2mm,
    left=2mm,
    right=2mm,
    top=1mm,
    bottom=1mm,
    fonttitle=\bfseries,
    title=Update Prompt,
}

\newtcolorbox{deleteprompt}{
    colback=DeletePurple!10!white,
    colframe=DeletePurple,
    boxrule=0.6pt,
    arc=2mm,
    left=2mm,
    right=2mm,
    top=1mm,
    bottom=1mm,
    fonttitle=\bfseries,
    title=Delete Prompt,
}

\definecolor{lstbg}{HTML}{F7F7F9}
\definecolor{lstkw}{HTML}{1F6FB2}
\definecolor{lststr}{HTML}{B5651D}
\definecolor{lstcom}{HTML}{6A8E3F}
\definecolor{lstrule}{HTML}{D0D0D7}
\lstdefinestyle{trace}{%
    language=Python,
    basicstyle=\ttfamily\scriptsize,
    keywordstyle=\color{lstkw}\bfseries,
    stringstyle=\color{lststr},
    commentstyle=\color{lstcom}\itshape,
    xleftmargin=8pt,
    xrightmargin=4pt,
    breaklines=true,
    breakatwhitespace=false,
    columns=fullflexible,
    keepspaces=true,
    showstringspaces=false,
    upquote=true,
    extendedchars=true,
    inputencoding=utf8,
    literate={\$}{{\$}}1 {_}{{\_}}1 {✓}{{\checkmark}}1 {→}{{$\rightarrow$}}1
             {≥}{{$\geq$}}1 {≤}{{$\leq$}}1 {≠}{{$\neq$}}1
             {á}{{\'a}}1 {é}{{\'e}}1 {ü}{{\"u}}1,
    captionpos=b,
    aboveskip=4pt,
    belowskip=4pt,
}

\newtcolorbox{agentresponse}[1][]{%
  enhanced,
  colback=white, boxrule=0pt, frame hidden,
  arc=0pt, before skip=12pt, after skip=12pt,
  left=6pt, right=6pt, top=8pt, bottom=8pt,
  fontupper=\small, breakable, #1}

\definecolor{systemgray}{RGB}{247,248,250}
\definecolor{systemgrayborder}{RGB}{120,130,145}

\newtcolorbox{systempromptbox}{
    colback=systemgray,
    colframe=systemgrayborder,
    boxrule=0.6pt,
    arc=2mm,
    left=2mm,
    right=2mm,
    top=1mm,
    bottom=1mm,
    fonttitle=\bfseries,
    title=System Prompt
}

\title{BIM-Edit: Benchmarking Large Language Models for IFC-Based Building Information Modeling}

\author{
Bharathi Kannan Nithyanantham$^{1}$\thanks{Equal contribution}
\And
Clemens Kujat $^{1}$\footnotemark[1]
\And
Tobias Sesterhenn$^{2}$\footnotemark[1]
\And
Stefan Telgmann$^{1}$
\And
Ashwin Nedungadi$^{1}$
\And
Jörn Plönnigs$^{1}$
\And
Christian Bartelt$^{2}$ \\
\And
Stefan Lüdtke$^{1}$
\\
$^{1}$ University of Rostock \\
$^{2}$ Clausthal University of Technology \\
\texttt{bharathikannan.nithyanantham@uni-rostock.de}
}

\begin{document}

\maketitle
\setcounter{footnote}{0}

\begin{abstract}


Large language models (LLMs) are increasingly applied to computer-aided design (CAD) to generate design artifacts from textual instructions. In engineering practice, this requires more than creating new geometry, models must also understand existing scenes, edit them correctly, and preserve semantics and relations. However, many CAD benchmarks focus on creating new models rather than editing existing ones, and mostly evaluate geometric correctness. We introduce BIM-Edit, a benchmark for evaluating LLMs on natural-language editing of Building Information Models (BIM) represented in the Industry Foundation Classes (IFC) format. BIM provides a challenging testbed because building models encode geometry together with semantic and relational structure. BIM-Edit contains 324 editing tasks spanning 11 realistic building models and 36 synthetic scenes. Tasks are expressed using three instruction categories -- direct, spatial, and topological -- covering both explicit and scene-grounded edits. We evaluate outputs along three dimensions: geometric accuracy, semantic validity, and topological consistency. Across evaluated LLMs, the best-performing model achieves only $49.5\%$ average score across the three metrics, and no model fully solves more than $3.4\%$ of tasks. These results demonstrate a substantial gap between current LLM capabilities and the requirements of structured engineering design workflows.\footnote{Dataset available at: \url{https://huggingface.co/BIM-Edit}}

\end{abstract}

\section{Introduction}
\label{sec:introduction}


\begin{figure}
    \centering
    \includegraphics[width=1\linewidth]{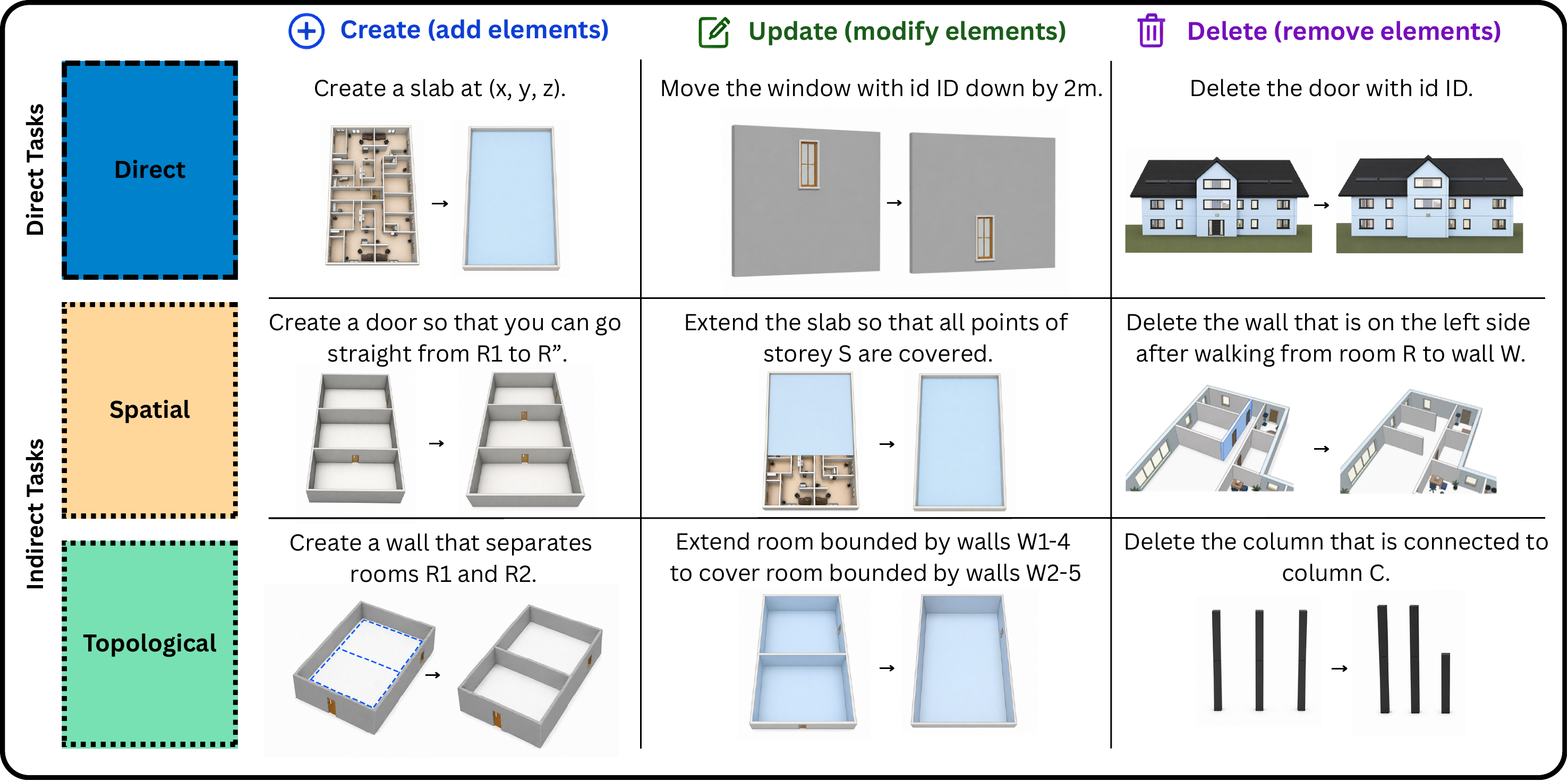}
    \caption{Overview of the tasks of BIM-Edit. BIM-Edit includes create, update, and delete tasks expressed as natural-language instructions of varying descriptions: direct (fully specified), spatial (context-based), and topological (relation-based).}
    \label{fig:intro}
\end{figure}

Large language models (LLMs) show strong performance on code generation~\citep{chen2024survey, gu2025effectiveness,jain2024livecodebench,jimenez2023swe}, which has motivated their use in 3D and computer-aided design (CAD). Recent systems can generate CAD sequences from textual descriptions \citep{khan2024text2cad, li2025cad, wang2025text} and from images of target models~\citep{chen2025cadcrafter, giannone2026gift}. However, most CAD benchmarks evaluate LLMs on small synthetic examples and require models to generate geometries from scratch~\citep{zhang2026large, du2024blenderllm, li2024llm4cad}. This setup diverges from real-world engineering practice, where CAD models are not purely geometric but also encode relationships between components and semantic properties under domain-specific constraints~\citep{qureshi2012statistical, corrado2022recent, wang2024framework, beitz1996engineering}. Additionally, in practice, CAD workflows rarely start from scratch and instead involve collaboration among experts on shared models~\citep{martins2013multidisciplinary}.

We identify three key discrepancies between current CAD benchmarks and real-world requirements. \textbf{(G1) Interaction with existing CAD models.}
A realistic benchmark should evaluate whether LLMs can interpret and modify existing CAD models.  Prior work has mostly focused on visual scene understanding in multimodal LLMs~\citep{azuma2022scanqa, liu2025can, ma20253dsrbench, zhan2025open3dvqa}. In contrast, the ability to retrieve information from and manipulate large CAD models via code remains largely underexplored~\citep{kienle2025querycad}. \textbf{(G2) Consistency of modifications.} Benchmarks should ensure that modifications to CAD models not only produce correct geometry but also preserve consistency in topology and semantics, such as element classes, material properties, containment, and connectivity. A model can be invalid from engineering perspective, even when it appears geometrically correct. For example, two structural beams that appear visually in contact, but are disjoint, compromise the structural integrity ~\citep{ploennigs2025building}. \textbf{(G3) Handling implicit natural language.} In real-world scenarios, edit requests are often underspecified and refer to elements through spatial, semantic, or topological relations. For example: \emph{``Enlarge the central hole of the wooden element''} only provides contextual spatial and semantic cues, but no explicit element IDs~\citep{yuan2025cad}.

BIM-Edit addresses those gaps: It operates on existing IFC models (G1) and evaluates LLMs in two complementary dimensions that target G2 and G3. The first dimension is valid BIM manipulation in create, update, and delete tasks. Each edit is scored on geometry, topology, and semantics metrics evaluating also engineering integrity (G2). The second dimension is contextual scene understanding. Each task is written in one of three instruction variants: direct, spatial, or topological, with increasing underspecification of prompts (G3). Direct instructions explicitly specify the target element and the required change. Spatial prompts specify the target indirectly through geometric context, such as position, direction, or distance. Topological prompts specify the target indirectly through BIM relations, such as adjacency, containment, hosting, or connectivity. These variants are used in both small models and larger realistic house models, allowing controlled evaluation across instruction ambiguity and scene complexity (see Figure~\ref{fig:intro}). The resulting benchmark is challenging for LLMs: The best of seven proprietary and open-weight models only reaches 49.48\% performance. Our contributions are:
\begin{itemize}
\item \textbf{BIM-Edit}, a benchmark of 324 natural-language editing tasks spanning create, update, and delete operations under direct, spatial, and topological instructions for small to large models.
\item A three-axis evaluation protocol that scores edits on geometric accuracy, semantic validity, and topological consistency rather than geometry alone.
\item We conduct experiments on seven recent LLMs to assess their performance on BIM-Edit and analyze their strengths and limitations in BIM editing.
\end{itemize}

\section{Related Work}

The related work can be grouped into benchmarks: (i) that more generally evaluate 3D scenes; (ii) that study CAD modeling in different applications; (iii) that investigate specifically BIM workflows.

\paragraph{3D scene understanding benchmarks.}
Existing work evaluates whether LLMs and MLLMs can reason about 3D scenes, for example through visual question answering or spatial reasoning tasks over object locations, relations, and scene composition \citep{azuma2022scanqa,liu2025can,ma20253dsrbench,zhan2025open3dvqa}. 
Mostly relevant in our context is emerging work on programmatic access to CAD artifacts, such as QueryCAD~\citep{kienle2025querycad}, which evaluates whether LLMs can synthesize code to extract information from CAD models. 
These benchmarks capture important parts of the indirect reasoning required in our setting, especially for spatial and relational references. However, they evaluate perception and reasoning in isolation: models are not required to generate executable CAD or BIM artifacts, nor to preserve consistency across geometry, semantics, and topology. 



\paragraph{CAD modeling benchmarks.}
Most CAD modeling benchmarks study \emph{generation} of complete models from modalities such as text~\citep{khan2024text2cad, li2025cad, wang2025text}, images~\citep{chen2025cadcrafter, giannone2026gift}, or other geometric inputs~\citep{wang20252d, seff2021vitruvion,dupont2024transcad}, often using large repositories such as ShapeNet~\citep{chang2015shapenet}, DeepCAD~\citep{wu2021deepcad}, ABC~\citep{koch2019abc}, and Fusion360~\citep{willis2021fusion}. For example, Text2CAD~\citep{khan2024text2cad} benchmarks text-to-CAD generation across prompts with different levels of specificity, which is conceptually related to our distinction between direct and indirect task formulations. Other work extends this setting to multimodal prompting or LLM-based evaluation of generated CAD models~\citep{du2024blenderllm,li2024llm4cad}. However, these benchmarks primarily assess model synthesis from scratch rather than modification of an existing artifact and, therefore, require no understanding of existing scenes. More closely related is a parallel line of work on \emph{CAD editing}, which targets natural-language edits over parametric CAD models: the model receives a source model and a natural-language edit instruction and must transform it into the desired target~\citep{yuan2025cad, hasanscope}. CAD-Editor~\citep{yuan2025cad} introduced this setting with a benchmark of create, update, and delete operations on existing CAD artifacts. A related setting is studied in BlenderGym~\citep{gu2025blendergym}, which considers edits from start-goal pairs, but assumes access to an explicit BlenderPython construction sequence. However, these benchmarks operate on CAD representations that primarily capture geometry, limiting their ability to assess engineering validity. In contrast, BIM models encode explicit relationships and domain semantics, enabling evaluation of structural consistency and functional correctness required for real-world Architecture, Engineering, and Construction (AEC) applications.


\paragraph{Building Information Modeling.}
Building Information Modeling is a life-cycle data-management paradigm that is established in civil engineering to ensure tool interoperability. It extends traditional geometry models in CAD by a semantic object-oriented model, with classes, attributes, and explicit relationships. The established data exchange format for this are Industry Foundation Classes (IFC), an open, vendor-neutral standard~\citep{iso16739_2024}, making it independent of proprietary BIM tools and formats (e.\,g., Revit, Vectorworks). 
Within IFC, building components (e.\,g., walls, slabs, windows) are defined as typed objects that encapsulate geometry, placement, properties, and inter-element relationships such as spatial containment. As a result, BIM models must maintain consistency of these properties, rather than being evaluated solely on geometric representation as in conventional CAD \citep{borrmann2018industry}.
Recent work has begun to explore LLM-based systems for BIM, targeting tasks such as information retrieval and Text2BIM generation \citep{hellin2025natural, du2024text2bim, zheng2023bim, deng2025bimgent, jang2024automated, fernandes2024gpt, liu2025bimcoder}. 
However, these approaches are typically validated on small from-scratch generation scenarios, and no unified benchmark systematically evaluates LLMs on BIM editing. BIM-Edit closes this gap with a broad cover of all create and edit tasks that test contextual understanding (G1), implicit prompting (G3), and result in valid engineering models (G2).

\section{Methodology}
\label{sec:methodology}


\begin{figure}[tbh]
    \centering
    \includegraphics[width=\linewidth]{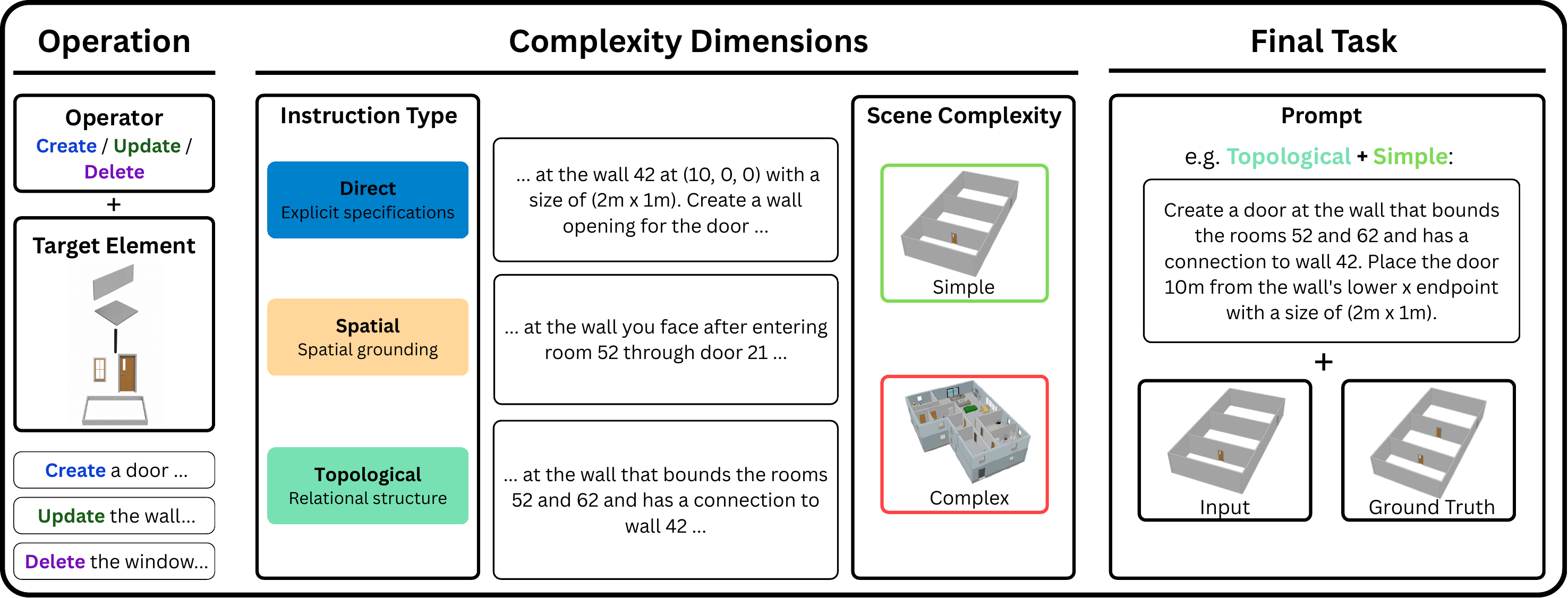}
    \caption{Overview of the task structure of the benchmark. Each task consists of an operator, a target element, and a combination of an instruction type and a scene complexity. Therefore, each operation can be done in six different ways (Instruction Type $\times$ Scene Complexity).}
    \label{fig:benchmark-tasks}
\end{figure}

BIM-Edit evaluates how well LLMs perform in editing existing structured 3D building models from natural-language instructions. The model must identify the referenced scene, apply the requested change, and preserve the rest of the model. This reflects real BIM workflows, where edits occur inside large shared models, and a visually plausible result can still be invalid if it breaks element types, spatial relations, or properties. Therefore, BIM-Edit evaluates the full edited model, not only its geometry. We define each BIM-Edit task as a triplet $(M^0, x, M^*)$, where $M^0$ is the input IFC model, $x$ is a natural-language edit instruction, and $M^*$ is the manually authored ground-truth model. The agent receives $(M^0, x)$ and must produce an edited model $M'$.
During the evaluation, $M'$ is then compared to $M^*$ with respect to geometric, semantic, and topological correctness.
This makes BIM-Edit independent of the agent design, since any system that reads and writes IFC files can be evaluated under the same protocol.

\subsection{Benchmark construction}

\paragraph{Modular Task Syntax.} BIM-Edit contains 324 tasks across 11 distinct large IFC models and 36 synthetic models. To enable reliable failure analysis, we structure the benchmark around tasks with controlled variations, as shown in Figure~\ref{fig:benchmark-tasks}.
Each prompt follows a strict pattern: (i) \textit{Operator} (create, update, delete), which specifies the underlying editing objective; (ii) \textit{Target Element} (e.\,g., wall, window, room); and (iii) \textit{Instruction}, which determines whether implicit information must be inferred from context. Instructions always have three variants and can refer to elements either directly (e.\,g., via element IDs) or indirectly through spatial or topological descriptions.
Additionally, tasks are evaluated in two \textit{scenarios}: (i) a simple scene with a limited number of elements, and (ii) a complex, realistic building, enabling assessment of the impact of increasing scene complexity.



This design allows to systematically evaluate tasks across different complexity dimensions from operator complexity (create, update, delete), to reference complexity (direct, spatial, topological), to scene complexity (simple, complex). In detail:

\paragraph{Operation complexity.} This dimension evaluates which operator types are handled well by the LLMs. \emph{Create} tasks require adding one or more new BIM elements. \emph{Update} tasks require modifying existing elements, for example by changing their size, position, or shape. Lastly, \emph{Delete} tasks require removing target elements and cleaning up dependent structure where necessary. We focus on common architectural entities as target elements, including walls, slabs, doors, windows, columns, and spaces. Furthermore, although the tasks always refer to exactly one target element, this does not mean that only one BIM element needs to be edited. For example, moving a window requires editing both the position of the window and the wall opening that it fills.
Spaces are a special case of building elements, as they represent non-physical, semantic constructs that are mostly defined by surrounding building elements such as walls, rather than physical objects themselves. As a result, modifying spaces may require indirect changes to adjacent structural elements.

\paragraph{Reference complexity.} 
Varying the instruction types assesses the contextual understanding and reasoning capabilities of the LLMs with respect to implicit user prompts (G3). 
First, \emph{direct} instructions explicitly name the target element (e.\,g., via ID) and fully specify all required parameters, including geometric properties and any relevant relations, thus requiring minimal scene understanding. Indirect instructions require the system to understand the scene by querying the IFC model first. Here, \emph{spatial} instructions use geometric context such as relative position, distance, orientation, or viewpoint, whereas \emph{topological} tasks rely on BIM relations including adjacency, containment, hosting, or connectivity. 
This uses the representational structure of IFC models, as they can be understood as semantic property graphs in which building elements form nodes with semantic classes and geometric representations, while relations between elements form edges. 
Consequently, the instruction categories induce different reasoning requirements: direct instructions rely on explicit element references, spatial instructions require geometric reasoning over the scene layout, and topological instructions require reasoning over relations such as adjacency, containment, and connectivity.

\paragraph{Scene complexity.} This dimension evaluates how well the LLMs can scale to large scenes providing potentially ambiguous context. We distinguish between \emph{simple} and \emph{complex} models based on structural complexity. 
Simple models consist of small BIM substructures with a limited number of elements and relations, designed to isolate specific editing behaviors. On average, they contain 21.03 elements and 98.74 relations. The complex models represent realistic house-like IFC files with larger layouts and denser relational structure. On average, they contain 614.88 elements and 2088.88 relations. All IFC models in BIM-Edit are manually created by human experts (G1). 
Examples of both models are shown in Appendix~\ref{app:ifc-examples}.
This distinction allows us to evaluate performance both in controlled settings and in realistic scenarios where the target element must be identified within a large and complex context, each setting containing $162$ tasks, respectively.

\subsection{Evaluation Metrics}
\label{sec:metrics}

We design a custom evaluation suite for BIM-Edit because standard CAD metrics mainly focus on geometry~\citep{zhang2026large}. A valid IFC edit must also be correct in its element definitions and topological relationships. For example, a prediction may match the reference shape exactly but still be partially invalid if it uses the wrong element class or adds a door without properly attaching it to its host wall. We therefore evaluate each prediction along three axes: geometry, semantics, and topology. The final score is defined as their unweighted mean:

\begin{equation}
S = \tfrac{1}{3}\bigl(S_{\text{geo}} + S_{\text{sem}} + S_{\text{topo}}\bigr),
\label{eq:final-score}
\end{equation}

where $S_{\text{geo}}$, $S_{\text{sem}}$, and $S_{\text{topo}}$ represent the geometry, semantic, and topology scores, respectively. Each score is bounded between 0 and 1, where higher is better.

Most edits affect only a small region of a much larger building model. As stated above, IFC models are semantic property graphs $G=(N,R)$, where building elements correspond to nodes $n \in N$ with semantic classes $c_n$, attribute sets $a_n$, and geometric representations $m_n$, while inter-element relations are edges $r \in R$ with a class $c_r$. Comparing complete IFC files would be dominated by unchanged graph structure. To correctly reflect edits, we evaluate only the edit graph $G_{\Delta}$. Let $G^*$ and $G'$ denote the semantic property graphs corresponding to the ground-truth model $M^*$ and the predicted model $M'$, respectively. Using a lightweight IFC diff built on IfcOpenShell~\citep{ifcopenshell}, we compute the edit graph $G_\Delta=(N_\Delta,R_\Delta)$ that identifies nodes and relations that were added, removed, or modified between the two graphs. Modifications do include also changes to semantic classes $c_n$, attributes $a_n$, geometry $m_n$, or relational types $c_r$. All metrics are computed on these edit sets, so the score reflects the quality of the edit itself rather than the unchanged parts of the model.
Together, the three metrics evaluate not only geometric correctness but also engineering validity by verifying that the semantic and topological consistency of the IFC model is preserved (G2).

\paragraph{Geometry Score.}
The geometry score measures whether the edit produces the correct shape in the correct location. We uniformly sample points from the surfaces for all $n \in N_{\Delta}$ in their reference $m^*_n$ and predicted geometry $m'_n$ to obtain point clouds $P^*$ and $P'$. We then compare these point clouds using the median bidirectional Chamfer distance $\mathrm{CD}{\mathrm{med}}$~\citep{fan2017point}. This set-level comparison supports edits that affect multiple objects without requiring one-to-one object matching. Since Chamfer distance is unbounded, we normalize it by the diagonal $D$ of the joint bounding box of the two point clouds and map it to $[0,1]$ using exponential decay:

\begin{equation}
S_{\text{geo}} = \exp\!\left(-\,\frac{\mathrm{CD}{\mathrm{med}}(P^*, P')}{D}\cdot\,\alpha\right),
\label{eq:geometry-score}
\end{equation}

where $S_{\text{geo}} = 1$ indicates a perfect geometric match, and $S_{\text{geo}} \to 0$ as the normalized geometric error increases. The parameter $\alpha=5$ controls how quickly the score decreases as the error increases.

\paragraph{Semantic Score.}

The semantic score measures whether the edited objects carry the correct type (G2). A geometrically valid result can still be incorrect if, for example, the agent represents a door as a window object. When node identities differ between $N’$ and $N^*$, we compute a one-to-one matching using the Hungarian algorithm~\citep{kuhn1955hungarian} based on oriented bounding box IoU (OBB-IoU). For each matched pair, we compute two semantic terms. The first term checks whether the predicted object has the same IFC class as the reference object. The second term measures the fraction of semantic properties relevant to the task that match, such as predefined type, name, material, or other attributes used by the task. For delete tasks, where no edited object remains to compare, the semantic terms is considered correct if the intended target object is successfully removed. The semantic score for a matched pair is the average of these two terms. The task-level semantic score is the mean over all reference edited objects after matching each reference object to its assigned prediction, with unmatched references assigned zero:

\begin{equation}
S_{\text{sem}} =
\frac{1}{|N_\Delta|}
\sum_{n \in N_\Delta}
\frac{1}{2}
\Bigl(
\mathbbm{1}\!\left[
c'_n =
c_n^*
\right]
+
\rho(a'_n, a_n^*)
\Bigr).
\label{eq:semantic-score}
\end{equation}

where $c_n$ denotes the IFC class of an object, $\mathbbm{1}$ is the identity operator, and $\rho(a'_n, a_n^*) \in [0,1]$ is the property score defined above.

\paragraph{Topology Score.}

The topology score measures changes in the graph edges, which are critical for IFC validity, for example, a missing connection on a load-bearing column can have serious structural implications. The metric directly evaluates changes on the difference graph $G_\Delta=(N_\Delta,R_\Delta)$ considering node and relation edits. To scale to large IFC graphs, we align node identities using a greedy bipartite matching heuristic and match relations only between aligned node pairs. We compute $\operatorname{F1}$ scores separately for edited nodes and edited relations, and combine them as


\begin{equation}
S_{\text{topo}} =
\lambda\,\operatorname{F1}(N', N^*) + (1-\lambda)\,\operatorname{F1}(R', R^*).
\label{eq:topology-score}
\end{equation}

We set $\lambda = 0.3$, so the relationship correctness receives the higher weight $(1-\lambda = 0.7)$ because the relations more strongly determine the validity of the model. If the reference edit contains no topological modifications, we assign $S_{\mathrm{topo}}=1$ when the prediction also introduces none, and $S_{\mathrm{topo}}=0$ otherwise.

\section{Evaluation}
\label{sec:evaluation}

\subsection{Experimental Setup} \label{sec:exp-setup}
We evaluate LLM agents in a code-execution environment for IFC editing. For each of the 324 tasks, an agent receives the natural-language instruction and the path to an input IFC model. The agent has access to a single tool, \textsc{execute\_ifc\_code(code: str)} , which executes generated Python code in a sandbox environment where the IFC model is preloaded using IfcOpenShell (see Appendix \ref{app:tool-description}). The agent must generate Python code and pass it to the tool to interact with the IFC model. It can use multiple tool calls, first to inspect the IFC model and later to apply edits such as creating geometry, changing placements, or editing semantic properties. For indirect instructions, the target elements and required operations are not explicitly stated. The LLM therefore has to query the IFC model, identify the relevant context, and ground its edit in spatial or relational reasoning. We adopt code generation rather than a curated toolset to provide a broad and flexible action space. This design evaluates whether LLMs can transfer general-purpose code generation capabilities to structured BIM editing without task-specific fine-tuning. A run ends when the agent declares the task is complete or reaches the budget of 20 tool calls. The final saved IFC file is the only artifact passed to the evaluator. 
We evaluate seven models on BIM-Edit, including proprietary and open-weight models: Gemini 3.0 Flash, Qwen 3.6 Plus, Claude Sonnet 4.6, GPT-5.4 Pro, GPT-5.4 Mini, Gemma 4 31B, and DeepSeek V3.2. All models are tested with the same agent harness, system prompt \ref{app:system-prompt}, and code-execution tool \ref{app:tool-description}. 

\subsection{Main Results}
\label{sec:results}


\paragraph{BIM-Edit exposes substantial gaps in current LLM capabilities.}
Table~\ref{tab:main-results-avg} summarizes overall performance on BIM-Edit. No evaluated model achieves an average score above $50\%$, highlighting the difficulty of reliable IFC-based BIM editing. Gemini 3.0 Flash achieves the best overall performance, followed by Qwen 3.6 Plus and Claude Sonnet 4.6. The per-metric breakdown reveals complementary model strengths: Gemini 3.0 Flash achieves the highest geometry and semantic scores, whereas Qwen 3.6 Plus performs best on topology. Across all models, geometry scores are consistently higher than semantic and topology scores, suggesting that current LLM agents can often approximate the correct shape while failing to preserve IFC semantics and relational consistency. The large standard deviations across all metrics further indicate substantial task-level variance across edit operations, instruction types, and scene contexts. The different rankings across metrics support our choice to evaluate BIM edits separately for geometry, semantics, and topology. 



\begin{table}[t]
\centering
\small
\caption{%
Average BIM-Edit scores across 324 tasks on a 0 to 100 scale.
Final is the average of geometry, semantics, and topology metrics.
Entries are mean {\color{black!55}\scriptsize$\pm$ std} across tasks.
}
\label{tab:main-results-avg}
\renewcommand{\arraystretch}{1.1}
\setlength{\tabcolsep}{4pt}
\newcommand{\sd}[1]{\,{\color{black!55}\scriptsize$\pm$#1}}
\begin{tabular}{lcccc}
\toprule
\textbf{Model}
& \textbf{Final} $\uparrow$
& \textbf{Geom.} $\uparrow$
& \textbf{Sem.} $\uparrow$
& \textbf{Topo.} $\uparrow$ \\
\midrule
Gemini 3.0 Flash   & \textbf{49.48}\sd{32.72} & \textbf{68.87}\sd{43.55} & \textbf{41.81}\sd{46.67} & 37.77\sd{38.46} \\
Qwen 3.6 Plus      & 47.63\sd{35.35}          & 59.21\sd{46.87}          & 35.93\sd{44.14}          & \textbf{47.77}\sd{42.65} \\
Claude Sonnet 4.6  & 45.31\sd{34.72}          & 55.90\sd{47.95}          & 33.85\sd{44.10}          & 46.19\sd{43.66} \\
GPT-5.4 Pro        & 43.94\sd{34.35}          & 50.14\sd{47.86}          & 38.69\sd{44.64}          & 42.97\sd{40.89} \\
DeepSeek V3.2      & 43.21\sd{35.19}          & 50.97\sd{47.30}          & 34.80\sd{44.41}          & 43.86\sd{42.73} \\
GPT-5.4 Mini       & 39.79\sd{34.21}          & 45.18\sd{47.29}          & 38.80\sd{45.89}          & 35.38\sd{39.46} \\
Gemma 4 31B        & 37.54\sd{36.17}          & 44.94\sd{48.07}          & 29.94\sd{44.61}          & 37.75\sd{41.94} \\
\bottomrule
\end{tabular}
\end{table}

\begin{table}[t]
\centering
\footnotesize
\caption{Strict BIM-Edit solve rates (\%) at a 98\% correctness threshold. \textit{Overall Solve Rate} reports tasks where geometry, semantic, and topology pass simultaneously: \textit{All} over the full 324-task benchmark, and \textit{Direct}, \textit{Spatial}, \textit{Topological} restricted to the corresponding 108-task instruction subsets. Evaluation Dimension reports the share of all 324 tasks that pass each metric individually.}
\label{tab:main-results-pass}
\renewcommand{\arraystretch}{1.05}
\setlength{\tabcolsep}{2.8pt}

\resizebox{\linewidth}{!}{%
\begin{tabular}{l cccc ccc}
\toprule
&
\multicolumn{4}{c}{\textbf{Overall Solve Rate (\%)}}
&
\multicolumn{3}{c}{\textbf{Evaluation Dimension (\%)}} \\
\cmidrule(lr){2-5}\cmidrule(lr){6-8}
\textbf{Model}
& \textbf{All} {\scriptsize(324)}
& \textbf{Direct} {\scriptsize(108)}
& \textbf{Spatial} {\scriptsize(108)}
& \textbf{Topological} {\scriptsize(108)}
& \textbf{Geom.}
& \textbf{Sem.}
& \textbf{Topo.} \\
\midrule
Gemini 3.0 Flash   & 1.5          & 0.93          & 1.85          & 1.85          & \textbf{40.7} & \textbf{35.8} & 14.5 \\
Qwen 3.6 Plus      & \textbf{3.4} & 4.63          & \textbf{3.70} & 1.85          & 37.3          & 25.9          & 27.8 \\
Claude Sonnet 4.6  & 1.9          & 2.78          & 0.93          & 1.85          & 36.7          & 25.9          & \textbf{29.9} \\
GPT-5.4 Pro        & 1.2          & 3.70          & 0.00          & 0.00          & 27.2          & 28.7          & 20.7 \\
DeepSeek V3.2      & 2.2          & \textbf{5.56} & 0.93          & 0.00          & 29.9          & 25.9          & 26.2 \\
GPT-5.4 Mini       & 0.6          & 1.85          & 0.00          & 0.00          & 23.8          & 31.8          & 16.0 \\
Gemma 4 31B        & 0.9          & 2.78          & 0.00          & 0.00          & 26.9          & 26.2          & 21.3 \\
\bottomrule
\end{tabular}%
}
\end{table}

\paragraph{Fully correct BIM edits remain extremely rare.}
While the average scores reflect the current capabilities of LLMs on the three evaluation dimensions, they do not directly reflect how many tasks are actually solved completely, i.e. achieving perfect scores on all metrics.
For this, we consider a task as solved, if for each metric the LLM achieves a score of at least $98\%$.
Table~\ref{tab:main-results-pass} shows that the models only solve the tasks for a very small proportion, with Qwen 3.6 Plus achieving the best solve rate of only $3.4\%$.
Models generally perform better on direct tasks than indirect tasks, which is especially true for the smaller models GPT-5.4 Mini and Gemma 4, which do not solve any spatial or topological task.
Analyzing the pass rate for each evaluation dimension, the best models solve around $40\%$ of tasks regarding the geometric task dimension but mostly below $30\%$ on the semantic and topological evaluation dimension.

\begin{figure*}[t]
    \centering

    \begin{subfigure}[t]{0.9\textwidth}
        \centering
        \includegraphics[width=\linewidth]{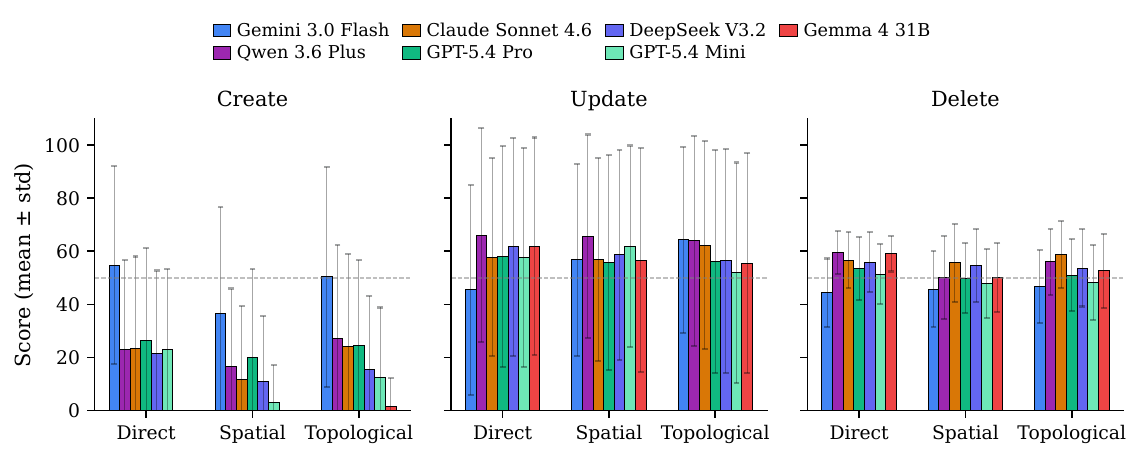}
        \caption{Performance by edit operation and instruction type.}
        \label{fig:direct-spatial-topological}
    \end{subfigure}

    \vspace{0.8em}

    \begin{subfigure}[t]{0.49\textwidth}
        \centering
        \includegraphics[width=\linewidth]{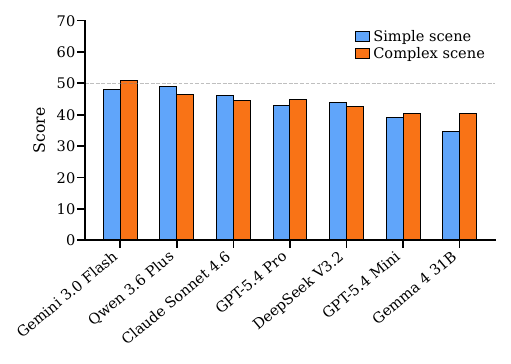}
        \caption{Performance by scene complexity.}
        \label{fig:artificial-realistic}
    \end{subfigure}
    \hfill
    \begin{subfigure}[t]{0.49\textwidth}
        \centering
        \includegraphics[width=\linewidth]{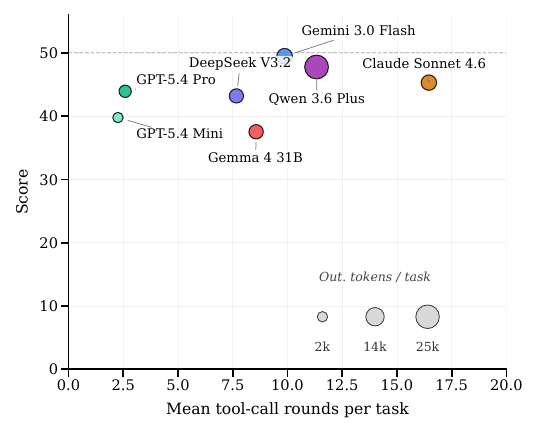}
        \caption{Performance versus mean tool-call rounds per task.}
        \label{fig:operational-effort-text}
    \end{subfigure}

    \caption{
    Additional BIM-Edit analyses.
    (a) Performance by edit operation and instruction category.
    (b) Comparison between simple and complex scenes.
    (c) Relationship between BIM-Edit score and agent interaction length.
    }
    \label{fig:additional-analysis}
\end{figure*}

\paragraph{Operation \& Reference complexity: Create operations are substantially harder.}
Performance differs strongly across both edit operations and instruction categories, as shown in Figure~\ref{fig:direct-spatial-topological}. 
Update tasks achieve the highest scores across all models, followed by delete tasks, whereas create tasks are consistently the most challenging setting. 
The additional difficulty arises because creation requires the agent not only to generate new geometry, but also to assign correct IFC semantics, place the element coherently within the scene, and establish valid relations to surrounding elements.
Interestingly, instruction type has only a limited effect on aggregated task scores. Performance remains similar across instruction categories within each edit operation, with indirect tasks occasionally even outperforming direct ones.
Combined with the strict solve rates in Table~\ref{tab:main-results-pass}, this indicates that models can often satisfy individual evaluation dimensions on indirect tasks, while failing to jointly preserve geometry, semantics, and topology in a fully correct BIM edit.
Furthermore, indirect tasks require substantially more output tokens. Averaged across all runs, direct instructions require $6.2k$ output tokens per task, compared to $11.4k$ and $9.9k$ for spatial and topological instructions, respectively. This suggests that models compensate for the additional reference ambiguity through longer reasoning and interaction traces, without consistently translating this additional computation into fully correct BIM edits.

\paragraph{Scene complexity: Larger IFC scenes do not substantially increase difficulty.}
To assess whether larger IFC scenes increase the difficulty of BIM editing, we compare performance across the two scene complexity settings in Figure~\ref{fig:artificial-realistic}. 
Although the complex IFC models contain substantially more elements and relations, most models achieve comparable scores across both settings. 
This suggests that model size alone is not the dominant source of difficulty in the experimental setup, likely because the IFC models are accessed programmatically rather than being placed directly into the LLM context, though this may become more relevant in future benchmark settings with different interaction paradigms.



\paragraph{Operational Effort: Longer agent trajectories do not guarantee better edits.}
Figure~\ref{fig:operational-effort-text} compares final BIM-Edit scores with the amount of agent interaction used during evaluation. 
This analysis is relevant because BIM-Edit requires iterative inspection and modification of structured IFC models, making longer interaction trajectories a potential indicator of more extensive reasoning or recovery behavior. 
However, the results show that longer trajectories do not necessarily translate into better performance. 
Qwen 3.6 Plus generates the most output tokens (25.5k), and Claude Sonnet 4.6 uses the largest number of tool-call rounds (16.5), yet Gemini 3.0 Flash achieves the highest overall score. 
In contrast, GPT-5.4 Pro and GPT-5.4 Mini rely on comparatively short interaction sequences while remaining competitive with several more verbose models. 


\subsection{Failure Cases}
\label{sec:failure}



\paragraph{Runtime failures are concentrated in distinct model-specific patterns.}
To better understand the operational robustness of LLM agents on BIM-Edit, Table~\ref{tab:efficiency-failures} reports the percentage of tasks ending in different runtime failure modes (definitions are provided in Appendix \ref{app:failure-modes-detail}). This analysis is important because BIM-Edit requires models not only to reason about IFC structures, but also to successfully manipulate them programmatically through IfcOpenShell. Without this distinction, benchmark performance could be dominated by failures in tool usage rather than limitations in BIM understanding itself. Since agents may already have modified the IFC model before termination, failed runs do not necessarily correspond to zero-score outputs.
The results reveal substantially different failure profiles across models. Claude Sonnet 4.6 exhibits the highest overall runtime failure rate (46.3\%), almost entirely caused by budget exhaustion (46.0\%), indicating that the agent frequently continues tool interaction until reaching the maximum tool-call limit. In contrast, DeepSeek V3.2 fails primarily through process crashes and streaming timeouts, suggesting lower execution stability despite avoiding budget exhaustion entirely. Qwen 3.6 Plus shows a more distributed failure profile across multiple categories, whereas Gemini 3.0 Flash and Gemma 4 31B remain comparatively stable apart from moderate rates of budget exhaustion. The GPT models achieve the highest operational stability, with almost no runtime failures across the benchmark.

\begin{table}[th]
\centering
\caption{Runtime failure rates (\%) across the different models.}
\label{tab:efficiency-failures}
\renewcommand{\arraystretch}{1.0}
\setlength{\tabcolsep}{4pt}
\begin{tabular}{l c c c c c c}
\toprule
\textbf{Model}
& \textbf{Budget}
& \textbf{Crash}
& \textbf{API}
& \textbf{Timeout}
& \textbf{Other}
& \textbf{Overall} \\
\midrule
Gemini 3.0 Flash   & 13.3 & 0.9 & 0.3 & 0.0  & 0.0 & 14.5 \\
Qwen 3.6 Plus      & 17.6 & 2.2 & 3.1 & 1.2  & 0.3 & 24.4 \\
Claude Sonnet 4.6  & 46.0 & 0.0 & 0.3 & 0.0  & 0.0 & 46.3 \\
GPT-5.4 Pro        & 0.0  & 0.0 & 0.6 & 0.0  & 0.0 & 0.6 \\
DeepSeek V3.2      & 0.0  & 5.9 & 2.5 & 12.3 & 0.0 & 20.7 \\
GPT-5.4 Mini       & 0.0  & 0.0 & 0.0 & 0.0  & 0.0 & 0.0 \\
Gemma 4 31B        & 8.3  & 4.6 & 1.9 & 0.0  & 0.0 & 14.8 \\
\bottomrule
\end{tabular}
\end{table}

\section{Conclusion \& Future Work}
\label{sec:conclusion}

We introduce BIM-Edit, a benchmark for natural-language editing of IFC-based building models that evaluates edits across geometric accuracy, semantic validity, and topological consistency. 
The benchmark contains 324 tasks spanning create, update, and delete operations under direct, spatial, and topological instructions on both simple substructures and complex building models. 
Across the seven evaluated LLMs, the best-performing model achieves an average score of only $49.48\%$, and no model fully solves more than $3.4\%$ of tasks.
The experiments show that current LLM agents can often make partial geometric progress, but rarely preserve geometry, semantics, and topology simultaneously. 
As a result, visually plausible edits frequently remain invalid as structured engineering artifacts. 
Create operations expose the largest capability gap, 
while, interestingly, the instruction type has only a limited effect on the aggregate task scores. 
However, indirect spatial and topological instructions substantially increase generation cost and are far less likely to result in fully correct BIM edits.
Similarly, increasing scene complexity has limited impact on overall performance in our setup, suggesting that the dominant bottleneck is not raw scene scale, but the ability to execute valid structured edits after the relevant IFC context has been identified. 
Overall, BIM-Edit shows that structured BIM editing remains far from solved for current LLM agents. 
Future work should therefore research on improving the agents' precision in BIM editing, improving the coupling of scene understanding and program execution, while also optimizing model cost and response time.
Future benchmark development could focus on multi-step BIM editing tasks spanning multiple interconnected elements, moving beyond isolated single-object modifications.

\bibliographystyle{plainnat}  
\bibliography{references}


\appendix

\newpage

\section{Limitations}
\label{sec:limitations}

BIM-Edit makes several choices that should be considered when interpreting the results. 
First of all, each task uses a single human-authored ground-truth IFC model. 
This makes scoring deterministic, but it may penalize valid alternative edits when an instruction is spatially or topologically ambiguous. 
However, we argue that useful systems for real world tasks should also understand the design intent of the user given the model context, and due to the continuous scoring, alternative edits may still reflect a good score.
Furthermore, the benchmark focuses on six common architectural element types: walls, slabs, spaces, doors, windows, and columns. 
Therefore, we do not evaluate the LLM capabilities on further relevent areas, such as Mechanical, Electrical, and Plumbing (MEP) systems, detailed structural systems, furniture, property-only edits, or multi-discipline coordination. 
Third, the agent harness in the conducted experiments is intentionally minimal and uses a single code-execution tool, without retrieval, schema-aware tools, or multi-agent orchestration. 
Therefore, the reported scores should be interpreted as out-of-the-box LLM performance that can be treated as a baseline instead of an upper bound of a production BIM assistant. 
Some runs are also limited by the 20-round tool-call budget, which may have an impact on the overall score.
However, in a real-world use case, the number of tool calls and output tokens represent a significant optimization objective, as acceptance to use such systems is directly influenced by the reliability as well as the response time of those systems.
The results of our work indicate that neither requirement is currently met, with models performing poorly while simultaneously incurring high interaction overhead through long execution traces.
The metric design choices, including the unweighted aggregation of geometry, semantics, and topology, the exponential normalization of geometric error, the $\lambda= 0.3$ weighting in the topology score, and the 98\% strict solve threshold, were selected to provide a consistent, interpretable, and computationally tractable evaluation across heterogeneous BIM edit types, while placing greater emphasis on relational correctness in IFC models. 
Although we do not provide a full sensitivity analysis or calibration against expert judgments, we expect the main conclusions to remain qualitatively stable, since performance gaps are large overall and the same trends appear consistently across the individual metrics.
Finally, the chosen metrics evaluate only the generated IFC artifact, not the generated code or reasoning traces. 
We therefore provide an analysis on example runs of GPT 5.4 Pro and Claude Sonnet 4.6 in Appendix~\ref{app:example-runs}.

\section{Broader Impact}
\label{sec:broader-impact}
AI-assisted BIM authoring has the potential to accelerate design work in architecture, engineering, and construction, but it also introduces risks: an incorrect edit in a safety-critical model, such as a load-bearing wall, a fire-rated partition, or a structural opening, can have consequences beyond those typical for software bugs. BIM-Edit is intended to help characterize these risks rather than to produce a fielded authoring assistant. We believe that output-level, structure-sensitive evaluation of the kind proposed here is a prerequisite for responsible deployment because it exposes failure modes that surface-level metrics miss. We release the benchmark under an open license and expect it to be used as an evaluation tool and not as a training signal for deployed authoring systems.
BIM-Edit is intended to expose limitations in critical engineering workflows. Therefore, high scores should not be interpreted as evidence that agents are ready for unsupervised use in safety-critical design tasks.

\section{Additional Related Work} \label{app:rel-work}

\paragraph{Agentic BIM approaches.}
Recent work has started to connect large language models to BIM workflows. \citet{zheng2023bim} study natural-language information retrieval over BIM data, with the goal of making model information easier to access without extensive manual interface engineering. \citet{hellin2025natural} propose an LLM-based workflow for natural-language querying over IFC-encoded BIM models. Another line of work moves from retrieval toward direct model modification. Text2BIM takes a further step toward authoring by using a multi-agent LLM framework to generate semantically rich BIM models from textual instructions inside a BIM authoring environment \citep{du2024text2bim}. Text2MBL similarly generates executable BIM code from text, but it focuses on modular building layouts rather than general IFC editing \citep{wei2025text2mbl}. 
These systems show that language-driven BIM interaction is feasible, but they do not provide a shared benchmark for artifact-level BIM editing. Their evaluations are typically tied to a specific authoring tool, task distribution, or workflow objective. Our benchmark differs in two main ways. First, it focuses on modifying an existing model rather than only generating a model from scratch. Second, it evaluates the final IFC artifact across geometry, topology, and semantics, rather than only assessing intermediate tool use or visual plausibility.
Lastly, \citet{yang2026far} proposed a benchmark for evaluating the physical plausibility of 3D house generation using Vision-Language Model (VLM) agents. Although their setting focuses on reconstructing houses from images, the benchmark is closely related to our work because it evaluates not only geometric reconstruction quality, but also physical constraints such as the structural and topological validity of generated IFC models.

\section{Benchmark Details and Dataset Card} \label{app:dataset}

\subsection{Purpose and Scope}

BIM-Edit was created to address a gap in the evaluation of LLMs for structured 3D building models. Existing CAD benchmarks mainly assess whether a system can generate plausible geometry \cite{guan2025cad, li2025cad}, but mostly they do not test whether object-level edits also preserve semantic properties and relational structure. In building design, these properties are tightly connected. A BIM file is correct only when its geometry is accurate, its elements have the appropriate BIM classes and attributes, and its relational graph encodes the required connections between elements. BIM-Edit formalizes this joint requirement as an evaluation target. The benchmark is intended for researchers working on AI for architecture, engineering, and construction, for developers evaluating IFC-capable LLMs, and for ablation studies that measure how instruction specificity affects structured editing performance.



\subsection{Scene Collection and Task Construction} \label{app:task-construction}

BIM-Edit contains 47 human-authored IFC models. These models are realistic building models and controlled substructures created by the authors to isolate specific element behaviors and reduce noise from unrelated model content. All models are released under CC-BY~4.0. 

Each task in BIM-Edit is defined as a triplet $(M^0, x, M^*)$, where $M^0$ is the input IFC model, $x$ is a natural-language edit instruction, and $M^*$ is a manually authored ground-truth model. The tasks were created using two techniques. In the first technique, the authors identified editable target elements that already existed in an IFC model and wrote a direct instruction for the corresponding edit. In the second technique, the authors manually applied a controlled edit to an IFC model, taken the resulting model as $M^*$, and then wrote an instruction describing that edit. The IFC models were edited using Blender with the Bonsai add-on and Revit.
For each task, the spatial and topological instruction variants were derived manually from the same underlying edit. This ensures that all three instruction variants refer to the same $M^*$. The design separates the effect of instruction type from task difficulty, because the target edit is identical across the three variants. Each task was verified by a second author to confirm that the ground truth was unambiguous and that the instruction fully specified the required edit.

\subsection{Task Statistics}

Table \ref{tab:task-stats} summarizes the distribution of tasks by element type and edit operation. BIM-Edit contains 324 tasks in total, split between human-authored scenes and controlled artificial scenes. The three instruction categories are balanced: direct, spatial, and topological instructions each contribute 108 tasks. It is not fully uniform across element types because some elements support more realistic edit variants than others. 

\begin{table}[t]
\centering
\caption{
Task distribution in BIM-Edit by element type and edit operation.
Counts are aggregated over all instruction categories and both scene types.
}
\label{tab:task-stats}
\small
\setlength{\tabcolsep}{9pt}
\begin{tabular}{lrrrr}
\toprule
\textbf{Element type} 
& \multicolumn{1}{c}{\textbf{Create}} 
& \multicolumn{1}{c}{\textbf{Update}} 
& \multicolumn{1}{c}{\textbf{Delete}} 
& \multicolumn{1}{c}{\textbf{Total}} \\
\midrule
IfcWall   & 24 & 24 & 24 & 72 \\
IfcSlab   & 24 & 24 & 24 & 72 \\
IfcSpace  & 24 & 24 & 24 & 72 \\
IfcDoor   & 18 &  6 & 12 & 36 \\
IfcWindow &  6 & 18 & 12 & 36 \\
IfcColumn & 12 & 12 & 12 & 36 \\
\midrule
\textbf{Total} & \textbf{108} & \textbf{108} & \textbf{108} & \textbf{324} \\
\bottomrule
\end{tabular}
\end{table}

\section{Full Evaluation Metric Definitions} 
\label{app:metrics}
BIM-Edit evaluates the changed parts of the 3D model instead of comparing full model files. This prevents scores from being dominated by unchanged building content. For each task, let $M^0$ be the input model, $M^*$ be the ground-truth model, and $M'$ be the model produced by the LLM agent. We construct a reference edit set $G_\Delta^*$ from $M^0$ and $M^*$, and a predicted edit set $G'_\Delta$ from $M^0$ and $M'$. Entities are marked as added, removed, or modified based on their class, identifier, geometry, and selected properties. These edit sets are then used to compute the geometry, semantic, and topology scores.

\paragraph{Per-task-type construction.}
The contents of $G_\Delta^*$ and $G'_\Delta$ depend on the task type. For \emph{create} tasks, $G_\Delta^*$ contains the ground-truth object to be created, while $G'_\Delta$ contains the entities added by the agent relative to $M^0$. For \emph{update} tasks, both sets contain the final state of the target object in $M^*$ and $M'$, respectively. Newly added entities are also included in $G'_\Delta$, so updates implemented by deleting and recreating the target can still be scored. For \emph{delete} tasks, $G_\Delta^*$ contains the entities that should be removed from $M^*$, while $G'_\Delta$ contains the entities actually removed by the agent. The diff uses the target ID specified in the task when available and falls back to an added-entity diff if the agent deletes and recreates the target under a new identifier.

\subsection{Geometry Score}

\paragraph{Aggregate point-cloud comparison.}
We uniformly sample points from the surface of each entity in the reference edit set and from each entity in the predicted edit set. The sampled points are combined into two edit-level point clouds: $P^*$ for the ground-truth edit and $P'$ for the agent-produced edit. Sampling is subject to a per-object budget of 4096 surface points and a total budget of 16384 points across all edited objects. The per-object allocation is proportional to surface area and is clamped at a minimum of 256 points per object. The headline geometry score is based on the median Chamfer distance between $P^*$ and $P'$:
\begin{equation}
S_{\text{geo}} =
\exp\!\left(
-\frac{\mathrm{CD}_{\mathrm{med}}(P^*, P')}{D}
\cdot 5.0
\right),
\end{equation}
where $\mathrm{CD}_{\mathrm{med}}(P^*, P')$ is the median bidirectional nearest-neighbor distance between the two point clouds, and $D$ is the diagonal length of the joint axis-aligned bounding box enclosing both point clouds. The distance is normalized by $D$ to make the score comparable across edits of different scales. The constant $5.0$ is a fixed decay factor that controls how quickly the score decreases as the normalized geometric error increases.

\paragraph{Additional geometry diagnostics.}
The evaluator additionally computes oriented bounding-box IoU (OBB-IoU), axis-aligned bounding-box IoU (AABB-IoU), voxel IoU at 0.05\,m resolution on a $128^3$ grid, F-score at $\tau = 0.01$ of the bounding-box diagonal, Jensen-Shannon divergence of voxel occupancy, and Hausdorff distance. These metrics are reported as additional diagnostics and do not contribute to the final score.

\subsection{Semantic Score}

The semantic score measures whether the edited entities carry the correct BIM meaning. Since semantic checks require object correspondences, we first match entities in the predicted edit set to entities in the reference edit set using oriented bounding-box IoU (OBB-IoU). The resulting OBB-IoU cost matrix is solved as a one-to-one assignment using the Hungarian algorithm~\citep{kuhn1955hungarian}. Matched predicted entities with an OBB-IoU below 0.05 against their assigned reference entity are discarded. Delete tasks are treated separately, since a successful edit leaves no predicted edited entity to compare against the reference entity. For these tasks, semantic correctness is defined by target removal: the score is 1.0 if the intended IFC entity is removed from the model, and 0.0 otherwise.

For each remaining matched pair $(n', n^*)$, the semantic score has two components:
\begin{enumerate}
\item Class score: The score is 1.0 if the class type of $c'_n$ matches the IFC class of $n^*_n$, and 0.0 otherwise. For example, a predicted IfcWall matched to a reference IfcWall receives 1.0, while a predicted IfcSlab or a proxy element matched to a reference IfcWall receives 0.0. 
\item Property score: The score is the fraction of task-relevant property keys whose values in $a'_n$ match the corresponding values in $n^*_n$ within a relative tolerance of 5\%. The properties \texttt{Tag}, \texttt{Description}, and \texttt{LongName} are excluded from this comparison.
\end{enumerate}

The per-pair semantic score is the average of the class score and the property score. The task-level semantic score is the mean over all reference entities, with unmatched reference entities assigned a score of zero.

\subsection{Topology Score}
We represent each IFC model as a typed property graph, where nodes correspond to structurally relevant IFC entities and edges correspond to standard IFC relations. These include spatial containment (\texttt{IfcRelContainedInSpatialStructure}), aggregation (\texttt{IfcRelAggregates}), space boundaries (\texttt{IfcRelSpaceBoundary}), opening and voiding (\texttt{IfcRelVoidsElement}), filling (\texttt{IfcRelFillsElement}), and element connection (\texttt{IfcRelConnectsElements}). We construct graphs for $M^0$, $M^*$, and $M'$. The reference topology edit is the symmetric difference between the graphs of $M^0$ and $M^*$, and the predicted topology edit is the symmetric difference between the graphs of $M^0$ and $M'$.

Because internal IFC identifiers, such as GUIDs and line numbers, may not be preserved across agent outputs, nodes between the predicted and reference edits must be aligned before comparison. Unlike the semantic score, where object correspondences directly determine the evaluated class and property terms and are therefore computed using an optimal Hungarian assignment, the topology score uses node matching only as an identity-canonicalization step before comparing relation deltas. Since topology graphs can contain hundreds of nodes and thousands of relations, we use a lightweight bipartite matching heuristic based on IFC class agreement and spatial proximity of local placements. We construct candidate correspondences, sort them by matching score, and greedily select non-conflicting node pairs. This avoids constructing dense assignment problems for large graphs while preserving the purpose of the topology metric, which is to evaluate whether the correct relational edits are present after alignment.

After alignment, we compute precision, recall, and F1 separately for node edits and edge edits. Node edits include added, removed, or modified entities, while edge edits include added or removed relations. The two F1 scores are combined with $\lambda = 0.3$ as in Eq.~\ref{eq:topology-score}, with edges weighted more heavily because relations determine whether an edited element is properly integrated into the building model. As a special case, when the reference edit requires no topology change, the score is $1$ if the prediction also introduces no topology change and $0$ otherwise. This penalizes prediction-only spurious topology edits.



\section{Experimental Details}
\label{app:exp-details}

\subsection{Hardware and Software Environment}

All inference is performed through cloud APIs. After inference, the evaluation pipeline itself runs on CPU only. On a standard desktop processor, evaluating a single task takes 1-2 minutes on average. Our parallelized evaluation script scores the full 324-task benchmark in under 2 hours once the inference outputs are available. The LLM pipeline and evaluator are implemented in Python~3.13 or later, with IfcOpenShell~0.8.x used as the IFC backend. All dependencies are pinned in the repository's requirements file.

\subsection{Model Versions and Configurations}

Table \ref{tab:model-slate} lists the models evaluated in this paper, together with their provider, checkpoint identifier, access route, and inference settings. All models are evaluated with the same system prompt, tool definition, tool-call budget, and evaluation pipeline. For Claude Sonnet 4.6, temperature and top-$p$ are left at the provider defaults because this model does not allow both parameters to be specified together. For all other models, temperature is set to 0 and top-$p$ is set to 1.0. The reasoning\_effort parameter is not overridden for any model, so each provider uses its default reasoning behavior.

\begin{table}[th]
\centering
\caption{
Model versions and inference configurations used in BIM-Edit.}
\label{tab:model-slate}
\footnotesize
\setlength{\tabcolsep}{4pt}
\begin{tabular}{lllcccc}
\toprule
\textbf{Model}
& \textbf{Provider}
& \textbf{Checkpoint ID}
& \textbf{Access}
& \textbf{Temp.}
& \textbf{Top-$p$}
& \textbf{Budget} \\
\midrule
Gemini 3.0 Flash  & Google        & gemini-3-flash-preview      & Native API & 0       & 1.0     & 20 \\
Qwen 3.6 Plus     & Alibaba Cloud & qwen/qwen3.6-plus           & OpenRouter & 0       & 1.0     & 20 \\
Claude Sonnet 4.6 & Anthropic     & claude-sonnet-4-6           & Native API & default & default & 20 \\
GPT-5.4 Pro       & OpenAI        & gpt-5.4-pro                 & Native API & 0       & 1.0     & 20 \\
DeepSeek V3.2     & DeepSeek      & deepseek/deepseek-v3.2      & OpenRouter & 0       & 1.0     & 20 \\
GPT-5.4 Mini      & OpenAI        & gpt-5.4-mini                & Native API & 0       & 1.0     & 20 \\
Gemma 4 31B       & Google        & google/gemma-4-31b-it       & OpenRouter & 0       & 1.0     & 20 \\
\bottomrule
\end{tabular}
\end{table}

\subsection{System Prompt}
\label{app:system-prompt}

All models receive the same system prompt:
\begin{systempromptbox}
\textit{You are a BIM assistant, with a deep knowledge in Building Information Modeling. You are working with IFC files and need to create IFCOpenshell calls to fulfill the task. Make sure to understand the current model before modification and modify the model always correctly on geometry, topology and semantics. Always understand the unit scale of the model. Use sensible defaults for unspecified values. Execute commands directly and never ask for confirmation.}
\end{systempromptbox}

\subsection{Tool Description}
\label{app:tool-description}

The agent is given a single tool: \texttt{execute\_ifc\_code(code:~str)~$\to$~str}. The tool is exposed to the model with the following description:
\begin{systempromptbox}
\textit{Execute Python code against the current IFC file. Input: \{code: str\}. The IFC model is pre-loaded as ifc (ifcopenshell.file). Also available: ifcopenshell, api (ifcopenshell.api), util (ifcopenshell.util), element\_util (ifcopenshell.util.element), and guid (ifcopenshell.guid). Assign to result to return data. Call commit() to save modifications.}
\end{systempromptbox} 
The input is a Python code string. The code runs in a sandboxed subprocess where the IFC file is pre-loaded into a variable using IfcOpenShell. The subprocess is restricted to the target IFC file and does not have access to unrelated files. Each tool call has a timeout of 120 seconds. If an inference call fails, the harness retries once after a 30-second delay, giving at most two inference attempts per task. After the agent completes or exhausts its tool-call budget of 20, the pipeline saves the current state of the IFC file to disk. This saved file is the only output passed to the evaluator.

\subsection{Reproducing Results}
\label{app:reproducing}

Every score reported in the main paper can be verified in the code folder that is included with the submission. Verification does not require rerunning model inference, because all cached outputs and evaluation results are provided. The released \texttt{runs/} directory contains the configuration files, cached outputs, and per-task evaluation reports for each of the seven evaluated models. For more information checkout the code repository.



\section{Additional Results and Analysis} \label{app:additional-results}

This section provides additional per-model breakdowns that support Section \ref{sec:results}.

\subsection{Failure-mode classifier}
\label{app:failure-modes-detail}

The counts in Table~\ref{tab:failure-modes} are computed deterministically from the per-task runtime metadata recorded in the cached run logs. Each task's first attempt is assigned to exactly one of six buckets:

\begin{itemize}
\item \emph{Executed}: the agent ran without raising any runtime error.
\item \emph{Budget exhausted}: the agent reached the 20-round tool-call budget, enforced as an internal limit of $20 \times 2 + 5 = 45$ graph nodes in langchain. 
\item \emph{Process crash}: the worker subprocess executing the task terminated unexpectedly.
\item \emph{API error}: the model provider returned a transport or protocol-level failure (HTTP error, malformed response, or read error from the client).
\item \emph{Streaming timeout}: a streaming chunk failed to arrive within the configured window.
\item \emph{Other runtime}: any remaining non-null error type.
\end{itemize}

\begin{table}[th]
\centering
\caption{Per-model runtime outcomes over 324 tasks. \emph{Executed} is the percentage of tasks the agent completed without a runtime error; the remaining columns count aborted runs by first failure type.}
\label{tab:failure-modes}
\renewcommand{\arraystretch}{1.15}
\begin{tabular}{l c c c c c c}
\toprule
\textbf{Model} & \textbf{Executed (\%)} $\uparrow$ & \textbf{Budget} & \textbf{Crash} & \textbf{API} & \textbf{Timeout} & \textbf{Other} \\
\midrule
GPT-5.4 Mini      & 100.00 &   0 &  0 &  0 &  0 & 0 \\
GPT-5.4 Pro       &  99.38 &   0 &  0 &  2 &  0 & 0 \\
Gemini 3.0 Flash  &  85.49 &  43 &  3 &  1 &  0 & 0 \\
Gemma 4 31B       &  85.19 &  27 & 15 &  6 &  0 & 0 \\
DeepSeek V3.2     &  79.32 &   0 & 19 &  8 & 40 & 0 \\
Qwen 3.6 Plus     &  75.62 &  57 &  7 & 10 &  4 & 1 \\
Claude Sonnet 4.6 &  53.70 & 149 &  0 &  1 &  0 & 0 \\
\bottomrule
\end{tabular}
\end{table}




\subsection{Task-Family Strengths and Weaknesses}
\label{app:task-strengths}

\begin{figure}[th]
    \centering
    \includegraphics[width=0.7\linewidth]{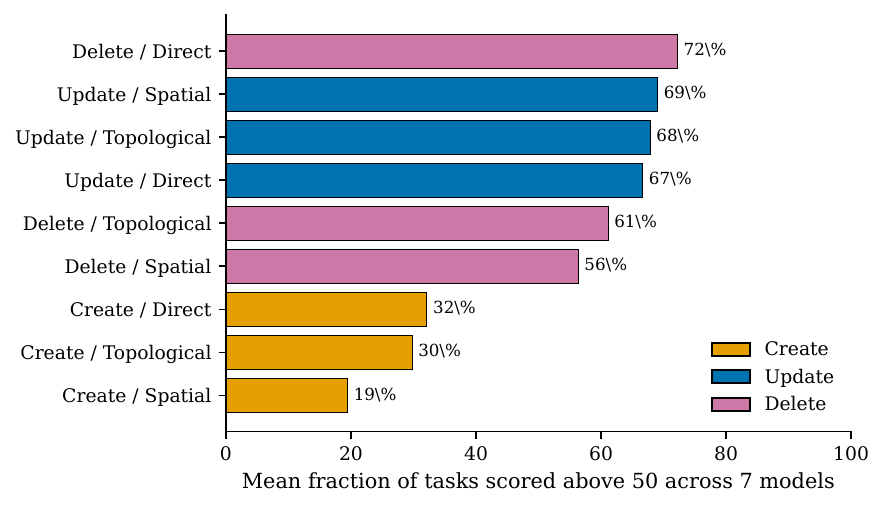}
    \caption{Mean fraction of tasks scoring above $50$ across the seven models.}
    \label{fig:task-strengths}
\end{figure}

Figure~\ref{fig:task-strengths} shows the fraction of tasks scoring above $50$ for each operation and instruction category, averaged across the seven models. 

\subsection{Metrics by Prompt Instruction Type}
\label{app:topology-by-instruction}

\begin{figure}[h!]
\centering
\includegraphics[width=\linewidth]{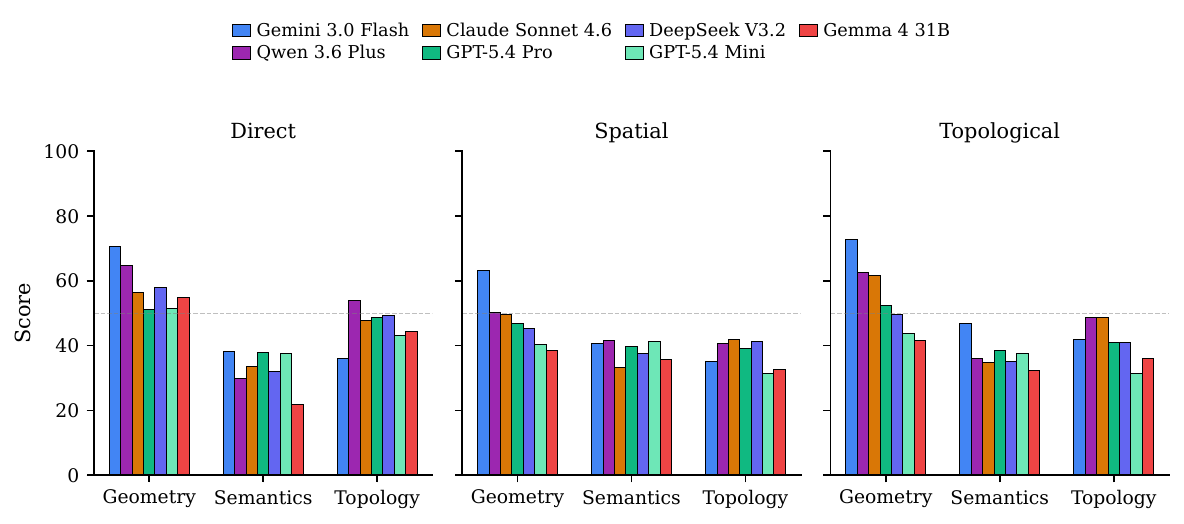}
\caption{Per-axis performance by instruction category across different metrics.}
\label{fig:topology-by-instruction}
\end{figure}

Figure~\ref{fig:topology-by-instruction} compares geometry, semantics, and topology scores across direct, spatial, and topological instructions. Each instruction category contains 108 tasks, so each plotted score is computed over the corresponding 108-task subset. Focusing on topology, direct instructions achieve the highest average score ($46.3$), followed by topological ($41.2$) and spatial instructions ($37.5$). Grouped together, the two indirect instruction types average $39.4$, about $7$ points below direct instructions. This suggests that topology performance is affected by how explicitly the relevant entities and relationships are given in the prompt. Topological instructions do not automatically lead to the best topology score because the instruction type and the evaluation metric capture different things: the instruction may use relational references to identify the target, while the metric checks whether the final IFC model satisfies the required relational structure. Direct instructions perform better because they expose more of the relational structure needed for IFC graph editing. We interpret this gap as evidence that current LLM agents benefit from explicit entity and relationship information.

\section{End-to-End Example Runs} \label{app:example-runs}

This section presents four complete agent runs in detail. For each run, we show the LLM-generated tool-call code, the agent's natural-language summary, and the per-axis BIM-Edit scores. The four examples are selected to cover the score range observed in the benchmark: a perfect run with all three axes scoring $1.0$, a high-partial run where two axes score $1.0$ and one axis drops, a run with incorrect geometric editing, and an all-zero run where the per-element matcher fails completely. 

The examples include two models, GPT-5.4~Pro and Claude~Sonnet~4.6. This makes it possible to compare their different agent behaviors. GPT-5.4~Pro often commits to a one-shot edit, while Claude~Sonnet~4.6 usually performs several read-only inspection rounds before applying the mutation. All four tasks come from the synthetic-scene subset of the benchmark, denoted as \emph{scene:~A} in the task tags. These scenes are small enough to inspect visually, which allows the edited IFC elements in the code to be cross-referenced with the rendered geometry. The four runs show a common pattern. The Python code emitted by the agent often looks good at first attempt, but small errors in the IFC relational structure can strongly reduce the geometry score. These errors can occur even when the agent's textual summary reports that the edit succeeded.

\subsection{Example~1: GPT-5.4~Pro on Door-Opening Translation (Successful)}
\label{app:ex-gpt54-door-success}

\paragraph{Task.} \texttt{DOR-UPD-DIR-A-001}.\;\emph{operation:} update,\;\emph{category:} direct,\;\emph{element:} \texttt{IfcDoor},\;\emph{scene:} A.
\begin{quote}
\small
move the door with the ID \texttt{3weZ9R4j52LPSdY\$3bzEMo}. Move the placement of its opening by 0.5\,m in $+x$-direction.
\end{quote}

\paragraph{Agent trace.} The agent solved the task in two tool-call rounds. Round~1 was discarded by the harness because it was a malformed call and returned \texttt{null}. In round~2, the agent retried with a small set of imports and applied the same edit. We show the round-2 code below, since this is the version that produced the saved IFC file.

\begin{lstlisting}[style=trace,caption={GPT-5.4~Pro tool call from round~2, which is the executed and committed round. The agent uses \texttt{FillsVoids} to find the opening, computes the project unit scale, and shifts the local placement coordinate of the opening's \texttt{IfcAxis2Placement3D}. The \texttt{commit()} call then writes the edited model back to disk.}]
import ifcopenshell
import ifcopenshell.util.unit

# Find door
door = ifc.by_guid('3weZ9R4j52LPSdY$3bzEMo')
if door is None:
    raise ValueError('Door with given GUID not found')

# Determine model unit scale (meters per model unit)
unit_scale = ifcopenshell.util.unit.calculate_unit_scale(ifc)
delta_model = 0.5 / unit_scale

# Find associated opening(s) through IfcRelFillsElement
openings = []
for rel in getattr(door, 'FillsVoids', []) or []:
    opening = rel.RelatingOpeningElement
    if opening:
        openings.append(opening)

if not openings:
    raise ValueError('No opening associated with the door via FillsVoids relationship')

moved = []
for opening in openings:
    placement = opening.ObjectPlacement
    if placement is None or not placement.is_a('IfcLocalPlacement'):
        raise ValueError(f'Opening {opening.GlobalId} has no IfcLocalPlacement')
    rel_place = placement.RelativePlacement
    if rel_place is None or not rel_place.is_a('IfcAxis2Placement3D'):
        raise ValueError(f'Opening {opening.GlobalId} placement is not IfcAxis2Placement3D')
    loc = rel_place.Location
    coords = list(loc.Coordinates)
    while len(coords) < 3:
        coords.append(0.0)
    old_coords = tuple(coords)
    coords[0] += delta_model
    loc.Coordinates = coords
    moved.append({
        'opening_guid': opening.GlobalId,
        'old_location': old_coords,
        'new_location': tuple(coords),
        'placement_relto_id': placement.PlacementRelTo.id() if placement.PlacementRelTo else None
    })

commit()
result = {
    'door_guid': door.GlobalId,
    'unit_scale_m_per_model_unit': unit_scale,
    'delta_applied_model_units': delta_model,
    'openings_moved': moved
}
\end{lstlisting}

\paragraph{Final response from the agent.}
\begin{agentresponse}
Done.\\
Moved the opening placement for the door \texttt{3weZ9R4j52LPSdY\$3bzEMo} by $+0.5$\,m in the X direction.\par\smallskip
\textbf{Details:}
\begin{itemize}\setlength\itemsep{1pt}
    \item Model unit scale: $1$\,m / unit
    \item Applied offset: $+0.5$ model units
    \item Opening moved: \texttt{1EHZix8UQ7pXxn8WcpugFE}
    \item Old opening location: $(2.5000,\, -0.0875,\, 0.0)$
    \item New opening location: $(3.0000,\, -0.0875,\, 0.0)$
\end{itemize}
The IFC file has been saved.
\end{agentresponse}

\paragraph{Score.} Table~\ref{tab:ex-gpt54-door-success} reports the per-axis BIM-Edit result for this run.
\begin{table}[!ht]
\centering
\small
\caption{Per-axis evaluation result for Example~1 (\texttt{DOR-UPD-DIR-A-001}, GPT-5.4~Pro).}
\label{tab:ex-gpt54-door-success}
\begin{tabular}{lcccccc}
\toprule
\textbf{Geometry} & \textbf{Semantics} & \textbf{Topology} & \textbf{Final} & \textbf{Tool calls} & \textbf{Out tokens} \\
\midrule
$1.0000$ & $1.0000$ & $1.0000$ & $\mathbf{1.0000}$ & $2$ & $1{,}191$ \\
\bottomrule
\end{tabular}
\end{table}

\begin{figure}
    \centering
    \includegraphics[width=\linewidth]{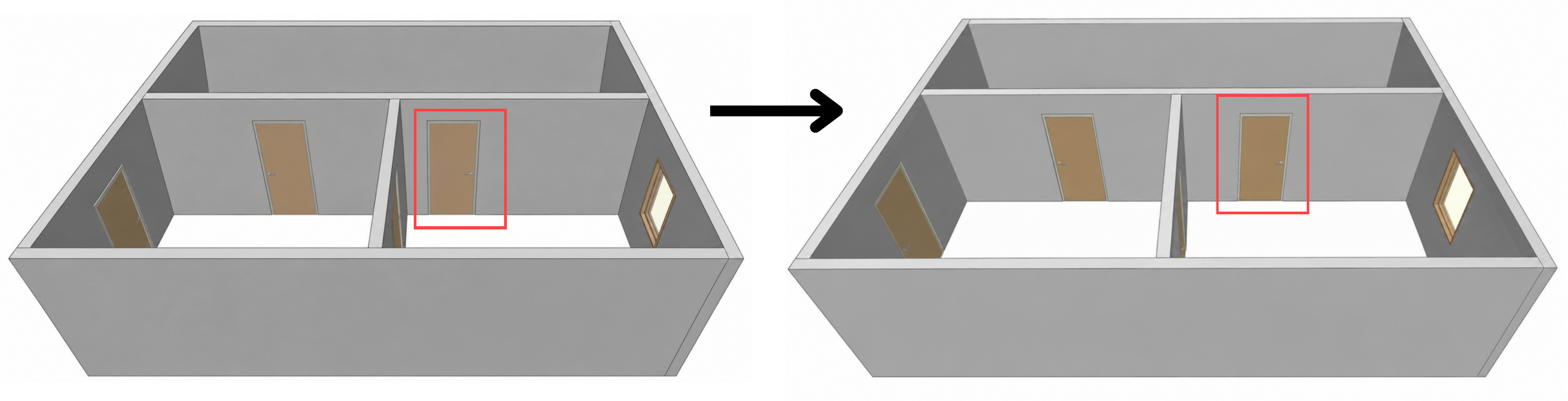}
    \caption{Input and edited IFC model.}
    \label{fig:ex1-vis}
\end{figure}

\paragraph{Visual Result.}
Figure~\ref{fig:ex1-vis} shows the input and edited IFC model.

\subsection{Example~2: Claude~Sonnet~4.6 on Wall Length Resize (High-Partial: $1.00/1.00/0.47$)}
\label{app:ex-sonnet-wall-partial}

\paragraph{Task.} \texttt{WAL-UPD-DIR-A-004}.\;\emph{operation:} update,\;\emph{category:} direct,\;\emph{element:} \texttt{IfcWall},\;\emph{scene:} A.
\begin{quote}
\small
change the length of the bounding box of the wall with the ID \texttt{0SW1L9fQP9xeGsED2XHlvo} to 10.852\,m, keeping its placement fixed.
\end{quote}

\paragraph{Agent trace (7 rounds, condensed).}
Sonnet used the first five rounds to inspect the wall, including its representation list, profile parameters (\texttt{XDim}, \texttt{YDim}, and \texttt{Depth}), axis polyline, placement, and project unit assignment. It found that the rectangular profile is centred at $(4.25625,\,0)$ with \texttt{RefDirection}~$=(-1,\,0)$, so the wall spans from local~$X{=}0$ to local~$X{=}8.5125$\,m. Round~6 is the only updating round. It updates three elements together: the profile \texttt{XDim}, the profile centre, and the axis polyline end point. Round~7 then verifies the result. 

\begin{lstlisting}[style=trace,caption={Round 6, the editing round. The agent updates \texttt{IfcRectangleProfileDef.XDim} to the new length, repositions the profile centre to $\texttt{new\_length}/2$ so the start end stays fixed, and updates the axis polyline end point. All three writes happen before \texttt{commit()}.}]
new_length = 10.852
wall = ifc.by_guid("0SW1L9fQP9xeGsED2XHlvo")

# Update Body representation: profile XDim and profile centre
for rep in wall.Representation.Representations:
    if rep.RepresentationIdentifier == "Body":
        for item in rep.Items:
            if item.is_a("IfcExtrudedAreaSolid"):
                profile = item.SweptArea
                profile.XDim = new_length
                profile.Position.Location.Coordinates = (new_length / 2, 0.0)

# Update Axis representation: polyline end point
for rep in wall.Representation.Representations:
    if rep.RepresentationIdentifier == "Axis":
        for item in rep.Items:
            if item.is_a("IfcPolyline"):
                # Points: first is start (0,0), second is end (length, 0)
                item.Points[1].Coordinates = (new_length, 0.0)

commit()
\end{lstlisting}

\paragraph{Final response from the agent.}
\begin{agentresponse}
The wall length has been successfully updated.\par\smallskip
\textbf{Changes made to wall \texttt{0SW1L9fQP9xeGsED2XHlvo}}:
\begin{center}
\small
\begin{tabular}{lcc}
\toprule
Property & Before & After \\
\midrule
Wall length (\texttt{XDim})  & $8.5125$\,m       & $\mathbf{10.852}$\,m \\
Profile centre $X$           & $4.25625$\,m      & $\mathbf{5.426}$\,m  \\
Axis end point               & $(8.5125,\, 0)$   & $\mathbf{(10.852,\, 0)}$ \\
\bottomrule
\end{tabular}
\end{center}
Kept fixed: \texttt{ObjectPlacement} at $(0.175,\, 4.0875,\, 0.0)$, wall thickness ($0.175$\,m), wall height ($2.8$\,m), extrusion direction (Z-axis).
\end{agentresponse}

\paragraph{Score.} Table~\ref{tab:ex-sonnet-wall-partial} reports the per-axis BIM-Edit result; geometry and semantics are perfect, topology is partial.
\begin{table}[!ht]
\centering
\small
\caption{Per-axis evaluation result for Example~2 (\texttt{WAL-UPD-DIR-A-004}, Claude~Sonnet~4.6). Geometry and semantics are perfect because the updated wall has the correct bounding box, \texttt{Ifc} class, and property set. Topology drops to $0.47$ because some connection relations to neighbouring walls were not re-checked after the wall length changed.}
\label{tab:ex-sonnet-wall-partial}
\begin{tabular}{lcccccc}
\toprule
\textbf{Geometry} & \textbf{Semantics} & \textbf{Topology} & \textbf{Final} & \textbf{Tool calls} & \textbf{Out tokens} \\
\midrule
$1.0000$ & $1.0000$ & $\mathbf{0.4714}$ & $0.8238$ & $7$ & $3{,}742$ \\
\bottomrule
\end{tabular}
\end{table}

\begin{figure}[h]
    \centering
    \includegraphics[width=\linewidth]{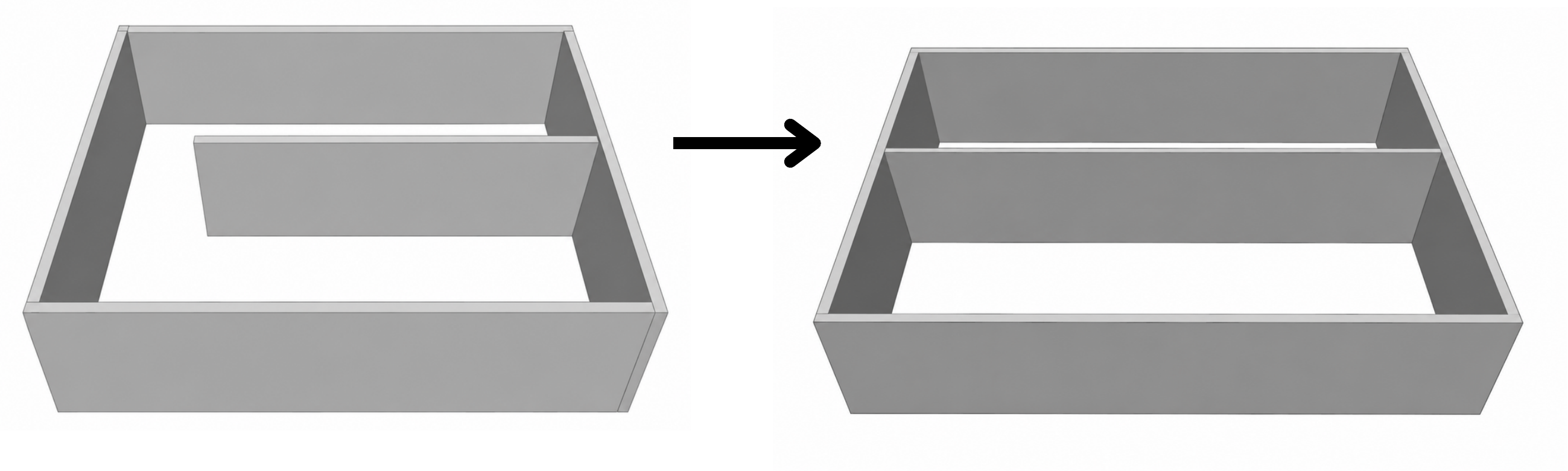}
    \caption{Input and edited IFC model.}
    \label{fig:ex2-vis}
\end{figure}

\paragraph{Visual Result.}
Figure~\ref{fig:ex2-vis} shows the input and edited IFC model.

\paragraph{Take-away.}
This is a high-partial outcome. The agent rebuilt the wall geometry correctly across the three coupled parameters: \texttt{XDim}, the profile centre, and the axis end point. As a result, the bounding box matches the ground truth. However, the final score does not reach $1.0$ because the wall was connected to neighboring walls through \texttt{IfcRelConnectsPathElements}. After resizing the wall, these connections were not revisited, leaving the topology axis at a partial value. This partial score is assigned when relations are present but do not match the ground-truth relation set. Overall, this example shows a common high-scoring update pattern: two axes are correct, but the third remains only partially valid.

\subsection{Example~3: GPT-5.4~Pro on Wall Thickness Resize (Low Geometry Score: $0.38/0.72/1.0$)}
\paragraph{Task.} \texttt{WAL-UPD-TOP-A-003}.\;\emph{operation:} update,\;\emph{category:} topological,\;\emph{element:} \texttt{IfcWall},\;\emph{scene:} A.
\begin{quote}
\small
change the wall thickness of the wall which is touching the walls with the IDs \texttt{0okaGdY4n1xulccRQajmzQ} and \texttt{0okaGdY4n1xulccRQajmzO} and is also bounding the rooms with the ID \texttt{2UsXAbj6n0fwEWqdkVuHvh} and \texttt{2f8KHsrajBsOOvPUvkuuAN} to 0.24m in $-x$-direction.
\end{quote}

\paragraph{Agent trace.} The agent solved the task in one tool-call round. It reasoned over connections to identify the target wall, adjusted the thickness property, and updated geometry. It also asjusted the material layer set thickness.

\begin{lstlisting}[style=trace,caption={The single tool call to solve the task.}]
import ifcopenshell
from ifcopenshell.util import element as element_util
from ifcopenshell.util import unit as unit_util
from ifcopenshell.util import placement as placement_util
from ifcopenshell.util import representation as repr_util

# Helpers
def get_global(elem):
    return getattr(elem, 'GlobalId', None)

def map_pts(op, fn):
    if op.is_a('IfcPolyline'):
        for p in op.Points:
            fn(p)
    elif op.is_a('IfcCompositeCurve'):
        for seg in op.Segments:
            parent = seg.ParentCurve
            map_pts(parent, fn)
    elif op.is_a('IfcTrimmedCurve'):
        basis = op.BasisCurve
        map_pts(basis, fn)
    elif op.is_a('IfcIndexedPolyCurve'):
        pts = op.Points.CoordList
        for i in range(len(pts)):
            fn((op.Points, i))
    else:
        pass


def get_axis2placement3d_matrix(pl):
    return placement_util.get_axis2placement(pl)


def local_point_from_world(mat, xyz):
    import math
    # mat is 4x4 nested tuple/list
    R = [[mat[r][c] for c in range(3)] for r in range(3)]
    t = [mat[r][3] for r in range(3)]
    v = [xyz[i] - t[i] for i in range(3)]
    # inverse for orthonormal rotation = transpose
    return (
        R[0][0]*v[0] + R[1][0]*v[1] + R[2][0]*v[2],
        R[0][1]*v[0] + R[1][1]*v[1] + R[2][1]*v[2],
        R[0][2]*v[0] + R[1][2]*v[1] + R[2][2]*v[2],
    )


def point_in_bbox_2d(pt, bbox, tol=1e-6):
    x, y = pt
    minx, miny, maxx, maxy = bbox
    return (minx-tol <= x <= maxx+tol) and (miny-tol <= y <= maxy+tol)


def bbox_intersects(a, b, tol=1e-6):
    return not (a[2] < b[0]-tol or b[2] < a[0]-tol or a[3] < b[1]-tol or b[3] < a[1]-tol)


def get_storey_of(elem):
    for rel in getattr(elem, 'ContainedInStructure', []) or []:
        if rel.RelatingStructure and rel.RelatingStructure.is_a('IfcBuildingStorey'):
            return rel.RelatingStructure
    return None

# Units
unit_scale = unit_util.calculate_unit_scale(ifc)

target_thickness_m = 0.24
target_thickness = target_thickness_m / unit_scale

# Identify reference walls and spaces
wall_a = ifc.by_guid('0okaGdY4n1xulccRQajmzQ')
wall_b = ifc.by_guid('0okaGdY4n1xulccRQajmzO')
space_a = ifc.by_guid('2UsXAbj6n0fwEWqdkVuHvh')
space_b = ifc.by_guid('2f8KHsrajBsOOvPUvkuuAN')

if not all([wall_a, wall_b, space_a, space_b]):
    result = {'error': 'One or more referenced elements not found'}
else:
    # Gather wall candidates on same storey, touching both reference walls and bounding both spaces
    storey = get_storey_of(space_a) or get_storey_of(space_b) or get_storey_of(wall_a) or get_storey_of(wall_b)
    candidates = []
    walls = ifc.by_type('IfcWall') + ifc.by_type('IfcWallStandardCase')
    seen = set()
    for w in walls:
        if w.id() in seen:
            continue
        seen.add(w.id())
        if w == wall_a or w == wall_b:
            continue
        if storey is not None and get_storey_of(w) != storey:
            continue
        # By boundaries to spaces
        bounded_spaces = set()
        for rel in ifc.by_type('IfcRelSpaceBoundary'):
            if rel.RelatedBuildingElement == w and rel.RelatingSpace:
                bounded_spaces.add(rel.RelatingSpace)
        if space_a not in bounded_spaces or space_b not in bounded_spaces:
            continue
        # Touching walls via connectivity if available
        connected = set()
        for rel in getattr(w, 'ConnectedTo', []) or []:
            if rel.RelatedElement: connected.add(rel.RelatedElement)
        for rel in getattr(w, 'ConnectedFrom', []) or []:
            if rel.RelatingElement: connected.add(rel.RelatingElement)
        if wall_a not in connected or wall_b not in connected:
            continue
        candidates.append(w)

    # Fallback: if no explicit connectivity, use space boundary and rough plan bbox touch tests
    if not candidates:
        def element_bbox_world_2d(elem):
            rep = getattr(elem, 'Representation', None)
            if not rep:
                return None
            mats = []
            try:
                m = placement_util.get_local_placement(elem.ObjectPlacement)
            except:
                return None
            minx = miny = float('inf')
            maxx = maxy = float('-inf')
            found = False
            for r in rep.Representations:
                items = list(r.Items)
                for it in items:
                    if it.is_a('IfcExtrudedAreaSolid'):
                        swept = it.SweptArea
                        pos = getattr(it, 'Position', None)
                        sm = placement_util.get_axis2placement(pos) if pos else ((1,0,0,0),(0,1,0,0),(0,0,1,0),(0,0,0,1))
                        # global = m * sm; for bbox of local profile points only translation/rotation needed
                        if swept.is_a('IfcRectangleProfileDef'):
                            xdim = swept.XDim
                            ydim = swept.YDim
                            pts = [(-xdim/2,-ydim/2,0),(xdim/2,-ydim/2,0),(xdim/2,ydim/2,0),(-xdim/2,ydim/2,0)]
                            pm = placement_util.get_axis2placement(swept.Position) if getattr(swept,'Position',None) else ((1,0,0,0),(0,1,0,0),(0,0,1,0),(0,0,0,1))
                            for x, y, z in pts:
                                # profile to swept
                                px = pm[0][0]*x + pm[0][1]*y + pm[0][3]
                                py = pm[1][0]*x + pm[1][1]*y + pm[1][3]
                                pz = pm[2][0]*x + pm[2][1]*y + pm[2][3]
                                # swept to object
                                ox = sm[0][0]*px + sm[0][1]*py + sm[0][2]*pz + sm[0][3]
                                oy = sm[1][0]*px + sm[1][1]*py + sm[1][2]*pz + sm[1][3]
                                # object to world
                                wx = m[0][0]*ox + m[0][1]*oy + m[0][3]
                                wy = m[1][0]*ox + m[1][1]*oy + m[1][3]
                                minx = min(minx, wx); miny = min(miny, wy)
                                maxx = max(maxx, wx); maxy = max(maxy, wy)
                                found = True
                        elif swept.is_a('IfcArbitraryClosedProfileDef') and swept.OuterCurve.is_a('IfcPolyline'):
                            for p in swept.OuterCurve.Points:
                                c = list(p.Coordinates) + [0]*(3-len(p.Coordinates))
                                px, py, pz = c[0], c[1], c[2]
                                ox = sm[0][0]*px + sm[0][1]*py + sm[0][2]*pz + sm[0][3]
                                oy = sm[1][0]*px + sm[1][1]*py + sm[1][2]*pz + sm[1][3]
                                wx = m[0][0]*ox + m[0][1]*oy + m[0][3]
                                wy = m[1][0]*ox + m[1][1]*oy + m[1][3]
                                minx = min(minx, wx); miny = min(miny, wy)
                                maxx = max(maxx, wx); maxy = max(maxy, wy)
                                found = True
            return (minx, miny, maxx, maxy) if found else None

        bbox_a = element_bbox_world_2d(wall_a)
        bbox_b = element_bbox_world_2d(wall_b)
        walls2 = []
        for w in walls:
            if w in (wall_a, wall_b):
                continue
            bounded_spaces = set()
            for rel in ifc.by_type('IfcRelSpaceBoundary'):
                if rel.RelatedBuildingElement == w and rel.RelatingSpace:
                    bounded_spaces.add(rel.RelatingSpace)
            if space_a not in bounded_spaces or space_b not in bounded_spaces:
                continue
            bw = element_bbox_world_2d(w)
            if bw and bbox_a and bbox_b and bbox_intersects(bw, bbox_a, tol=0.05/unit_scale) and bbox_intersects(bw, bbox_b, tol=0.05/unit_scale):
                walls2.append(w)
        candidates = walls2

    if len(candidates) != 1:
        result = {
            'error': 'Target wall not uniquely identified',
            'candidate_ids': [get_global(w) for w in candidates],
            'count': len(candidates)
        }
    else:
        wall = candidates[0]
        changed = {'wall': wall.GlobalId, 'changes': []}

        # Update material layer set usage / layer thickness if present
        assoc_rels = getattr(wall, 'HasAssociations', []) or []
        for rel in assoc_rels:
            mat = getattr(rel, 'RelatingMaterial', None)
            if not mat:
                continue
            if mat.is_a('IfcMaterialLayerSetUsage'):
                ls = mat.ForLayerSet
                if ls and len(ls.MaterialLayers) == 1:
                    old = ls.MaterialLayers[0].LayerThickness
                    ls.MaterialLayers[0].LayerThickness = target_thickness
                    changed['changes'].append({'type': 'material_layer', 'old': old, 'new': target_thickness})
            elif mat.is_a('IfcMaterialLayerSet'):
                if len(mat.MaterialLayers) == 1:
                    old = mat.MaterialLayers[0].LayerThickness
                    mat.MaterialLayers[0].LayerThickness = target_thickness
                    changed['changes'].append({'type': 'material_layer', 'old': old, 'new': target_thickness})

        # Update body geometry: move -X side so +X side stays fixed (thickness grows toward -x)
        rep = wall.Representation
        body_items = []
        if rep:
            for r in rep.Representations:
                if r.RepresentationIdentifier == 'Body' or r.RepresentationType in ('SweptSolid', 'Clipping', 'CSG', 'Brep'):
                    body_items.extend(list(r.Items))

        def adjust_profile(profile):
            # returns True if changed
            if profile.is_a('IfcRectangleProfileDef'):
                old = profile.XDim
                profile.XDim = target_thickness
                changed['changes'].append({'type': 'rectangle_xdim', 'old': old, 'new': target_thickness})
                # shift profile center by -(new-old)/2 in local x to extend toward -x only
                pos = profile.Position
                delta = (target_thickness - old) / 2.0
                old_loc = tuple(pos.Location.Coordinates)
                coords = list(old_loc) + [0.0]*(2-len(old_loc))
                coords[0] -= delta
                pos.Location.Coordinates = coords[:len(old_loc)] if len(old_loc) >= 2 else coords[:2]
                changed['changes'].append({'type': 'profile_position_shift_x', 'delta': -delta})
                return True
            elif profile.is_a('IfcArbitraryClosedProfileDef'):
                curve = profile.OuterCurve
                # only handle polyline / indexed polycurve by moving min-x points to new min-x
                pts = []
                if curve.is_a('IfcPolyline'):
                    for p in curve.Points:
                        c = list(p.Coordinates)
                        if len(c) < 2: continue
                        pts.append((p, c))
                    xs = [c[0] for _, c in pts]
                    if not xs:
                        return False
                    minx, maxx = min(xs), max(xs)
                    old = maxx - minx
                    dx = target_thickness - old
                    if abs(dx) < 1e-9:
                        return True
                    for p, c in pts:
                        if abs(c[0]-minx) < 1e-9:
                            c[0] -= dx
                            p.Coordinates = c
                    changed['changes'].append({'type': 'arbitrary_profile_minx_shift', 'old_thickness': old, 'new': target_thickness, 'delta': -dx})
                    return True
                elif curve.is_a('IfcIndexedPolyCurve'):
                    coordlist = curve.Points.CoordList
                    xs = [c[0] for c in coordlist if len(c) >= 2]
                    if not xs:
                        return False
                    minx, maxx = min(xs), max(xs)
                    old = maxx - minx
                    dx = target_thickness - old
                    if abs(dx) < 1e-9:
                        return True
                    newlist = []
                    for c in coordlist:
                        cc = list(c)
                        if len(cc) >= 2 and abs(cc[0]-minx) < 1e-9:
                            cc[0] -= dx
                        newlist.append(tuple(cc))
                    curve.Points.CoordList = newlist
                    changed['changes'].append({'type': 'indexed_profile_minx_shift', 'old_thickness': old, 'new': target_thickness, 'delta': -dx})
                    return True
            return False

        geom_changed = False
        for item in body_items:
            if item.is_a('IfcExtrudedAreaSolid'):
                geom_changed = adjust_profile(item.SweptArea) or geom_changed

        # Axis representation length unaffected, no change needed

        if not geom_changed and not changed['changes']:
            result = {'error': 'No editable thickness geometry/material found', 'wall': wall.GlobalId}
        else:
            commit()
            result = {
                'status': 'success',
                'unit_scale': unit_scale,
                'target_wall': wall.GlobalId,
                'target_thickness_model_units': target_thickness,
                'details': changed
            }
\end{lstlisting}

\paragraph{Final response from the agent.}
\begin{agentresponse}
Done. The wall was identified as:

\begin{itemize}
    \item 000irGlVjApOlI2apOJp9x
\end{itemize}

Applied changes:

\begin{itemize}
    \item Thickness set to 0.24 m
    \item Modified in -x direction
    \item Material layer thickness updated from 0.175 to 0.24
\end{itemize}

Note: the wall body used an IfcRectangleProfileDef, and its XDim was changed from 4.45 to 0.24 with a profile shift applied to keep the change toward -x.
\end{agentresponse}

\paragraph{Score.} Table~\ref{tab:ex-gpt-geom-low} reports the per-axis BIM-Edit result; every axis collapses to $0$.
\begin{table}[!ht]
\centering
\small
\caption{Per-axis evaluation result for Example~3 (\texttt{WAL-UPD-TOP-A-003}, GPT-5.4~Pro).}
\label{tab:ex-gpt-geom-low}
\begin{tabular}{lcccccc}
\toprule
\textbf{Geometry} & \textbf{Semantics} & \textbf{Topology} & \textbf{Final} & \textbf{Tool calls} & \textbf{Out tokens} \\
\midrule
$\mathbf{0.3813}$ & $\mathbf{0.7222}$ & $\mathbf{1.0000}$ & $\mathbf{0.7012}$ & $1$ & $3{,}701$ \\
\bottomrule
\end{tabular}
\end{table}

\begin{figure}[h]
    \centering
    \includegraphics[width=\linewidth]{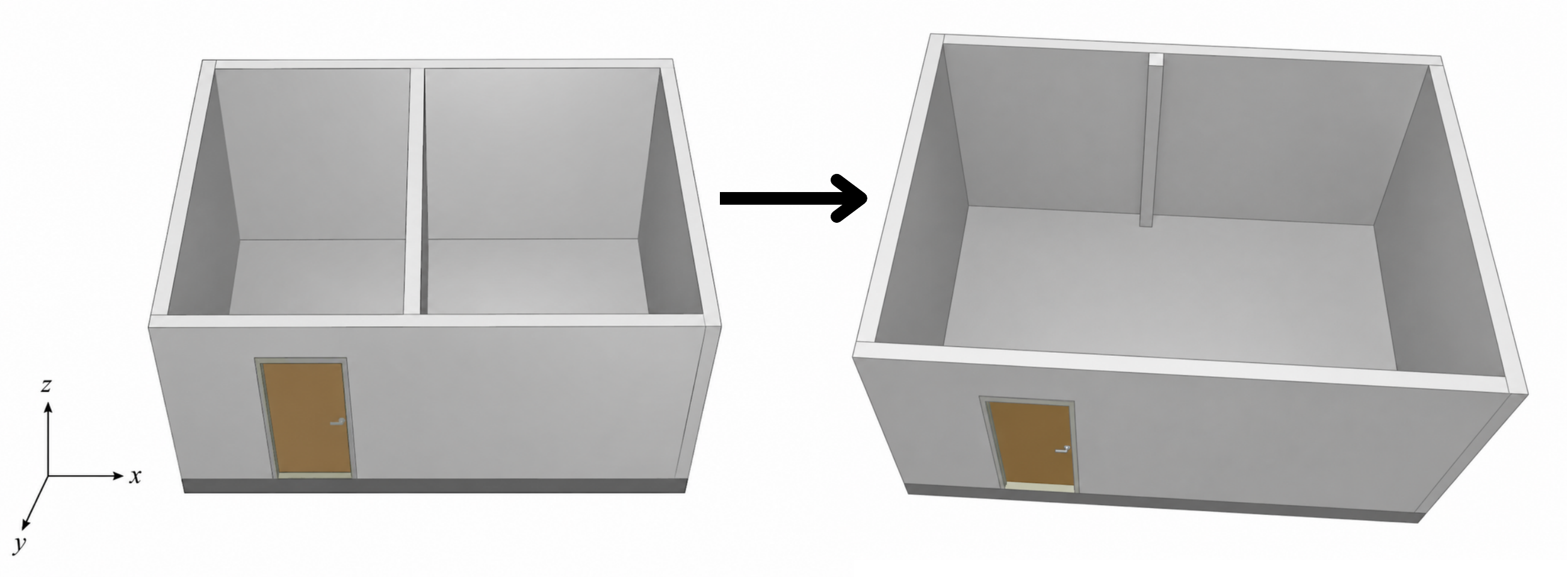}
    \caption{Input and edited IFC model.}
    \label{fig:ex3-vis}
\end{figure}

\paragraph{Visual Result.}
Figure~\ref{fig:ex3-vis} shows the input and edited IFC model.

\paragraph{Take-away.}
In BIM models, the keyword thickness typically refers to the y-direction by default. Thickness does not inherently correspond to the x-direction as initially intended; instead, the LLM relied on this conventional interpretation of the geometric property. The model therefore adjusted the thickness along the y-direction. However, since the wall is oriented orthogonally to the x-axis, its thickness should in this specific case be interpreted along the x-direction. In this sense, the error is semantic rather than purely geometric. Although the prompt explicitly specified an adjustment in the negative x-direction, the model still applied the change along the y-axis.

\subsection{Example~4: Claude~Sonnet~4.6 on Window Translation (All-Zero Catastrophic Failure)}
\label{app:ex-sonnet-window-zero}

\paragraph{Task.} \texttt{WIN-UPD-DIR-A-002}.\;\emph{operation:} update,\;\emph{category:} direct,\;\emph{element:} \texttt{IfcWindow},\;\emph{scene:} A.
\begin{quote}
\small
Move the window with ID \texttt{0okaGdY4n1xulccRQajpyj} by 1\,m in positive $y$ direction. Also move the opening \texttt{3H9tIOVCqvfsgJgyvP95X\$} the same and preserve all relationships.
\end{quote}

\paragraph{Agent trace (7 rounds, condensed).}
Rounds~1 to 5 are read-only inspections. The agent prints the placements for the window and the opening, checks the project unit assignment to confirm that the model is in metres, follows \texttt{IfcRelFillsElement} and \texttt{IfcRelVoidsElement} to identify the host wall, and finds the key invariant: the window's \texttt{PlacementRelTo} is the opening's placement. Round~6 is the only updating round, and round~7 prints the verification result.

\begin{lstlisting}[style=trace,caption={Round 6, the only updating round. }]
opening = ifc.by_guid("3H9tIOVCqvfsgJgyvP95X$")
op_location = opening.ObjectPlacement.RelativePlacement.Location

print(f"Before: {op_location.Coordinates}")     # (11.3, 0.1, 0.9)

current_coords = op_location.Coordinates
new_y = current_coords[1] + 1.0
op_location.Coordinates = (current_coords[0], new_y, current_coords[2])

print(f"After: {op_location.Coordinates}")      # (11.3, 1.1, 0.9)

# Verify the window's relative placement is unchanged
window = ifc.by_guid("0okaGdY4n1xulccRQajpyj")
win_ax = window.ObjectPlacement.RelativePlacement
assert window.ObjectPlacement.PlacementRelTo == opening.ObjectPlacement

commit()
\end{lstlisting}

\paragraph{Final response from the agent.}
\begin{agentresponse}
The operation is complete.\par\smallskip
\textbf{Movement applied.} Both the \emph{window} (\texttt{0okaGdY4n1xulccRQajpyj}) and the \emph{opening} (\texttt{3H9tIOVCqvfsgJgyvP95X\$}) were moved $+1.0$\,m in the positive $y$ direction.\par\smallskip
\textbf{How it was done.} The IFC placement hierarchy was leveraged: the opening's placement (\texttt{\#3577}) is shared as the \texttt{PlacementRelTo} for the window's placement, so updating only the opening's \texttt{IfcCartesianPoint} from $(11.3,\, 0.1,\, 0.9)$ to $(11.3,\, 1.1,\, 0.9)$ moves both elements together.\par\smallskip
\textbf{Relationships preserved.} \texttt{IfcRelFillsElement} (window fills the opening), \texttt{IfcRelVoidsElement} (opening voids the wall \texttt{0okaGdY4n1xulccRQajmzO}), and the placement hierarchy (window remains at $(0,0,0)$ relative to the opening) all unchanged.
\end{agentresponse}

\paragraph{Score.} Table~\ref{tab:ex-sonnet-window-zero} reports the per-axis BIM-Edit result; every axis collapses to $0$.
\begin{table}[!ht]
\centering
\small
\caption{Per-axis evaluation result for Example~4 (\texttt{WIN-UPD-DIR-A-002}, Claude~Sonnet~4.6). The agent's edit moves the window to coordinates that match the prompt, but those coordinates differ from the ground-truth window's location. So the per-element matcher fails and all axes are $0$.}
\label{tab:ex-sonnet-window-zero}
\begin{tabular}{lcccccc}
\toprule
\textbf{Geometry} & \textbf{Semantics} & \textbf{Topology} & \textbf{Final} & \textbf{Tool calls} & \textbf{Out tokens} \\
\midrule
$\mathbf{0.0000}$ & $\mathbf{0.0000}$ & $\mathbf{0.0000}$ & $\mathbf{0.0000}$ & $7$ & $3{,}262$ \\
\bottomrule
\end{tabular}
\end{table}

\begin{figure}[h]
    \centering
    \includegraphics[width=\linewidth]{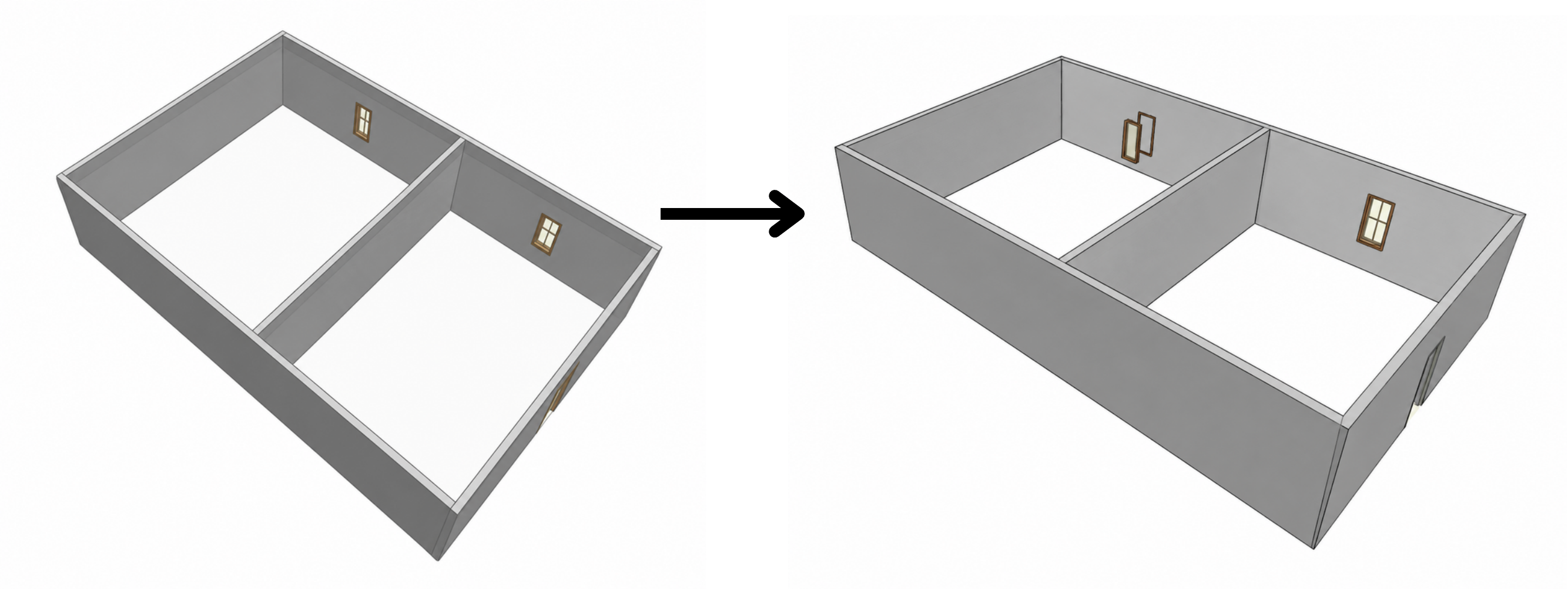}
    \caption{Input and edited IFC model.}
    \label{fig:ex4-vis}
\end{figure}

\paragraph{Visual Result.}
Figure~\ref{fig:ex4-vis} shows the input and edited IFC model.

\paragraph{Take-away}: Example~4 shows that a high iteration count can mainly reflect read-only work, such as element inspection, querying, and scene understanding. Even though they work, these steps do not prevent an all-zero outcome as the final edit is not correct. The intended edit needs to be correct to get high scores.

\section{Representative Task Examples}
\label{app:examples}

This section shows representative BIM-Edit tasks, grouped by element type and edit operation. All examples are taken from human-authored scenes. The dataset release includes the full set of 324 task prompts. Task IDs follow the pattern \texttt{ELEM-OP-INSTR-SCENE-NNN}, where \texttt{SCENE} is \texttt{R} for realistic scenes and \texttt{A} for artificial (synthetic) scenes.

\subsection{Wall Tasks}

\paragraph{Create task, direct instruction.}
\emph{Task ID: WAL-CRE-DIR-R-001.}
The direct create task specifies the wall geometry and the target relationship GUIDs explicitly. In this prompt, the agent must create the wall, assign it to the correct building storey, create the required wall-to-wall connections, and add the corresponding space-boundary relationships.
\begin{createprompt}
\small
\textit{Create a wall with a length of 2.62\,m from $(x_1,y_1,z_1){=}(7.180,\allowbreak 14.140,\allowbreak 3.000)$ to $(x_2,y_2,z_2){=}(9.800,\allowbreak 14.140,\allowbreak 3.000)$ with a thickness of 0.240\,m in $-y$ direction and a height of 3\,m. Make sure that the wall is added to the building storey with id 1jPCssxb5C1RSM\_YHIx\$zl, connected to walls 30a7nM35T5pgZzaItbPb1u and 153QDldl9AdhGy6O1dePed, and bounds the spaces 19QUlaWcT26g1KZHffEeW9 and 19QUlaWcT26g1KZHffEeWz.}
\end{createprompt}

\paragraph{Create task, spatial instruction.}
\emph{Task ID: WAL-CRE-SPA-R-001.}
The spatial create task specifies the wall position through scene context rather than explicit coordinates. In this prompt, the stairway is used as the spatial reference, and the agent must compute the wall position from the stairway geometry and the given offset.
\begin{createprompt}
\small
\textit{Create a wall 0.9\,m away in $-y$ direction from the stairway after going it up. Wall must have thickness of 0.240\,m in $y$ direction and a height of 3\,m.}
\end{createprompt}

\paragraph{Create task, topological instruction.}
\emph{Task ID: WAL-CRE-TOP-R-001.}
The topological create task specifies the wall through the relationship between two spaces rather than explicit coordinates or wall IDs. In this prompt, the agent must identify the two rooms, infer the boundary between them, and place the new wall along that boundary.
\begin{createprompt}
\small
\textit{Create a wall that separates the rooms (19QUlaWcT26g1KZHffEeW9) and (19QUlaWcT26g1KZHffEeWz) and has a height of 3\,m.}
\end{createprompt}

\paragraph{Update task, direct instruction.}
\emph{Task ID: WAL-UPD-DIR-R-001.}
\begin{updateprompt}
\small
\textit{Decrease the length of the wall with ID 30a7nM35T5pgZzaItbPb1c by 2.1\,m fixing one side at $x{=}0.2$. Delete the connecting relationship to wall 30a7nM35T5pgZzaItbPb1v. Make sure that the other relationships stay consistent.}
\end{updateprompt}

\paragraph{Update task, spatial instruction.}
\emph{Task ID: WAL-UPD-SPA-R-001.}
\begin{updateprompt}
\small
\textit{Decrease the length of the wall that is on the right side after entering room (19QUlaWcT26g1KZHffEeWX) through door (30a7nM35T5pgZzaItbPb1h) by 2.1\,m keeping the side with lower $x$ value fixed.}
\end{updateprompt}

\paragraph{Update task, topological instruction.}
\emph{Task ID: WAL-UPD-TOP-R-001.}
\begin{updateprompt}
\small
\textit{Decrease the length of the wall that bounds only the rooms (19QUlaWcT26g1KZHffEeWj) and (19QUlaWcT26g1KZHffEeWX) by 2.1\,m keeping the side with lower $x$ value fixed.}
\end{updateprompt}

\paragraph{Delete task, direct instruction.}
\emph{Task ID: WAL-DEL-DIR-R-001.}
\begin{deleteprompt}
\small
\textit{Delete the wall with ID 2AOGoBTWz3ieiE459aLhPN. Make sure to delete the wall from all its relationships: connections to walls 2AOGoBTWz3ieiE459aLeaP, 0FwbbEHzD6bwLn0l0aWD1y, and 0FwbbEHzD6bwLn0l0aWBr\_, bounding of spaces 19QUlaWcT26g1KZHffEedX, 19QUlaWcT26g1KZHffEedY, and 2KN4OKK7D6Uw\_xHkV6DTvG, and voided by openings 3p6vUUnefjnD\$Kee\_zq0zd and 0tUp8t\_xqOHBUun1YEiX\_v.}
\end{deleteprompt}

\paragraph{Delete task, spatial instruction.}
\emph{Task ID: WAL-DEL-SPA-R-001.}
\begin{deleteprompt}
\small
\textit{Delete the wall that is on the left side after walking from room (2KN4OKK7D6Uw\_xHkV6DTv6) to wall (2AOGoBTWz3ieiE459aLeaP).}
\end{deleteprompt}

\paragraph{Delete task, topological instruction.}
\emph{Task ID: WAL-DEL-TOP-R-001.}
\begin{deleteprompt}
\small
\textit{Delete the wall that bounds only the rooms (19QUlaWcT26g1KZHffEedX), (19QUlaWcT26g1KZHffEedY), and (2KN4OKK7D6Uw\_xHkV6DTvG).}
\end{deleteprompt}

\subsection{Door and Window Tasks}

Door and window tasks follow the same direct, spatial, and topological structure as the wall tasks. However, doors and windows involve additional IFC relationships, such as openings, voiding relationships, and filling relationships. The agent must place the element correctly and also satisfy these dependent relationships.

\paragraph{Door create task, topological instruction.}
\emph{Task ID: DOR-CRE-TOP-R-001.}
\begin{createprompt}
\small
\textit{Add a door to the wall that connects room 2gkzSyKgLDVALhNNC1fZ\_k to room 2gkzSyKgLDVALhNNC1fZ\_d. The door opening must have a distance of $-0.61$\,m to the wall edge with the maximum $y$ value, a width of 0.89\,m, and a height of 2.045\,m.}
\end{createprompt}

\paragraph{Window update task, spatial instruction.}
\emph{Task ID: WIN-UPD-SPA-R-001.}
\begin{updateprompt}
\small
\textit{Move the only window in the second storey that has no window right below down so that it aligns with the other windows of the new wall.}
\end{updateprompt}

\paragraph{Door delete task, topological instruction.}
\emph{Task ID: DOR-DEL-TOP-R-001.}
\begin{deleteprompt}
\small
\textit{Delete the door that connects the room 2gkzSyKgLDVALhNNC1fZ\_q to the outside.}
\end{deleteprompt}

\subsection{Slab Tasks}
Slab tasks focus on IFCSlab elements.

\paragraph{Slab create task, direct instruction.}
\emph{Task ID: SLB-CRE-DIR-R-001.}
\begin{createprompt}
\small
\textit{Create a slab starting at $(2.161,\allowbreak 4.310,\allowbreak -0.4)$ with a height of 0.2\,m in $z$ direction, a length of 30\,m in $x$ direction, and a width of 15\,m in $y$ direction. Assign it to the storey with ID 13LV5dTeP3CAox54x56C1Z.}
\end{createprompt}

\paragraph{Slab update task, spatial instruction.}
\emph{Task ID: SLB-UPD-SPA-R-001.}
\begin{updateprompt}
\small
\textit{Decrease the length of the slab below the ground storey by 7\,m keeping it fixed at the smallest $x$ value.}
\end{updateprompt}

\paragraph{Slab delete task, topological instruction.}
\emph{Task ID: SLB-DEL-TOP-R-001.}
\begin{deleteprompt}
\small
\textit{Delete the slab of the building storey with ID 13LV5dTeP3CAox54x56C1Z.}
\end{deleteprompt}

\subsection{Space Tasks}
Space tasks focus on IfcSpace elements, which represent rooms or usable areas in a building model. 

\paragraph{Space create task, direct instruction.}
\emph{Task ID: ROM-CRE-DIR-R-001.}
\begin{createprompt}
\small
\textit{Create a room at (0.438, 5.542, 0), with distances in x direction of 3.823m, in y direction of 4.399m, and in z direction of 2.438m. The area is bounded by walls 09aTDCDGbDbB5YMd6zhdRK, 2o7\$Px5Hf4\_OWCiwfVz3qS, 2o7\$Px5Hf4\_OWCiwfVz3\_g, and 2o7\$Px5Hf4\_OWCiwfVz3\$d. Add the according relationships to the walls, the slab 2o7\$Px5Hf4\_OWCiwfVz0eP and the building storey 13LV5dTeP3CAox54x56C1Z.}
\end{createprompt}

\paragraph{Space create task, spatial instruction.}
\emph{Task ID: ROM-CRE-SPA-R-001.}
\begin{createprompt}
\small
\textit{Create a room in the area that you can look into through the most central window of the wall with id 09aTDCDGbDbB5YMd6zhdRK. Make it 2.438\,m high.}
\end{createprompt}

\paragraph{Space create task, topological instruction.}
\emph{Task ID: ROM-CRE-TOP-R-001.}
\begin{createprompt}
\small
\textit{Create a room in the area that is enclosed by the walls 09aTDCDGbDbB5YMd6zhdRK, 2o7\$Px5Hf4\_OWCiwfVz3qS, 2o7\$Px5Hf4\_OWCiwfVz3\_g, and 2o7\$Px5Hf4\_OWCiwfVz3\$d with a height of 2.438\,m.}
\end{createprompt}

\subsection{IFC Models used in the Benchmark}
\label{app:ifc-examples}

Figure \ref{fig:main_4x3_grid} shows example IFC files used in this benchmark.

\newcommand{\ifccell}[1]{
    \begin{minipage}[c][3.3cm][c]{0.32\linewidth}
        \centering
        \includegraphics[width=\linewidth,height=3.3cm,keepaspectratio]{#1}
    \end{minipage}
}

\begin{figure*}[hbt!]
    \centering
    \ifccell{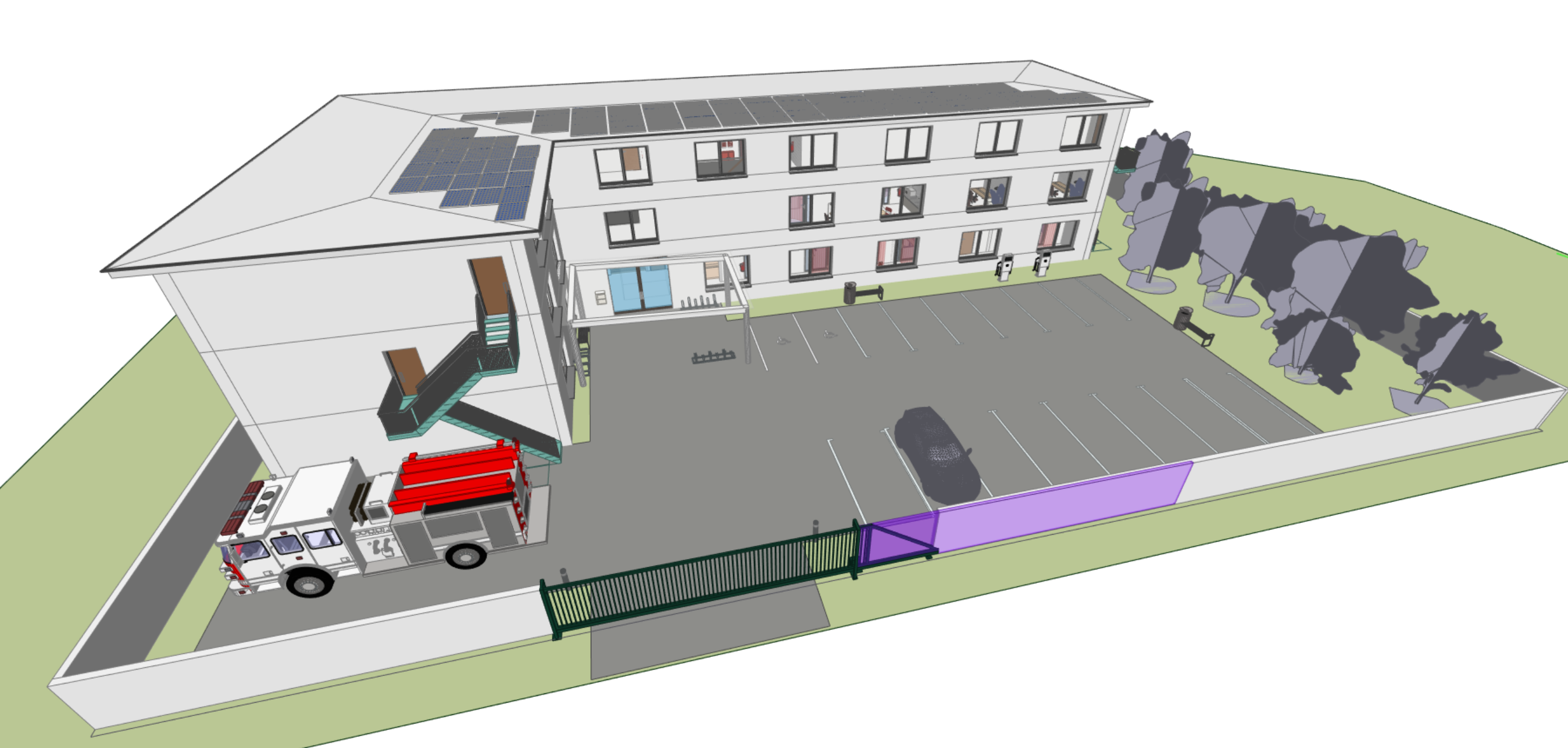}\hfill
    \ifccell{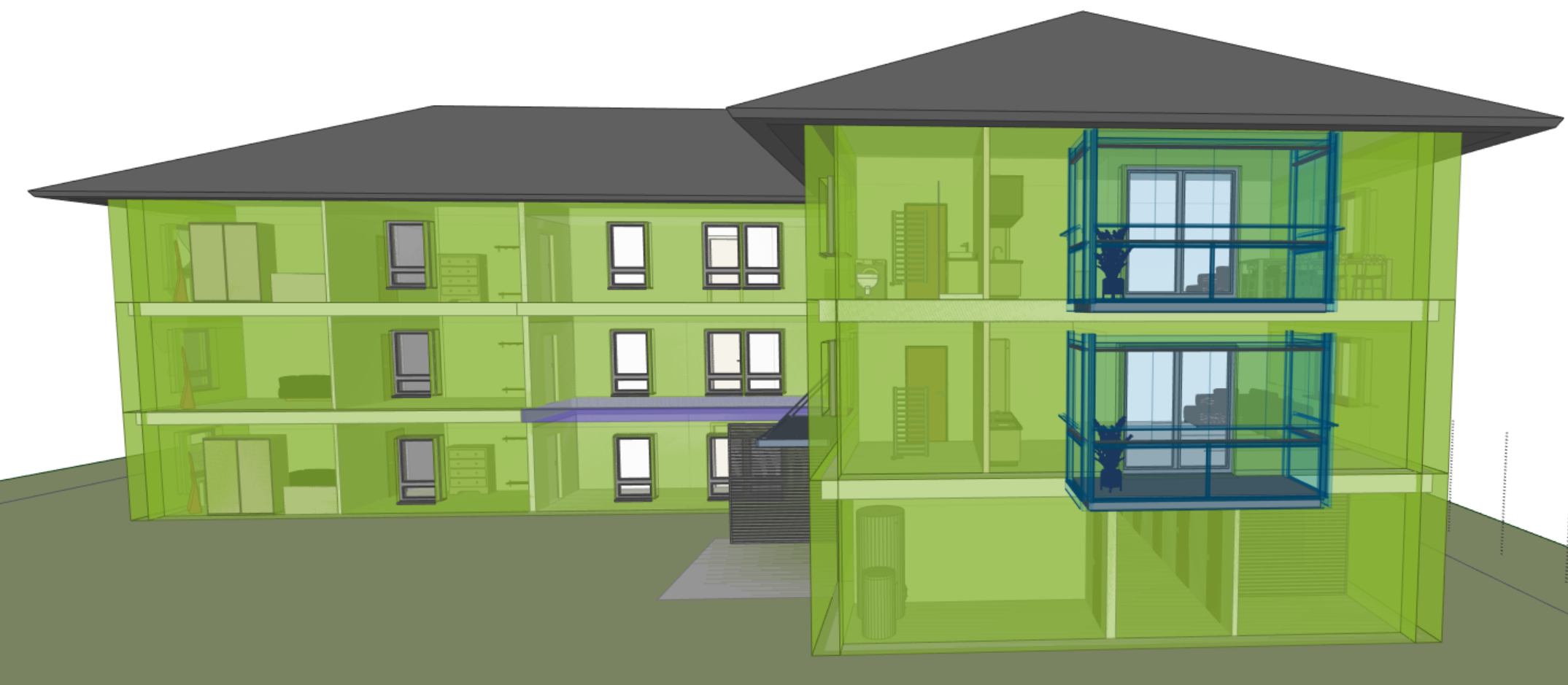}\hfill
    \ifccell{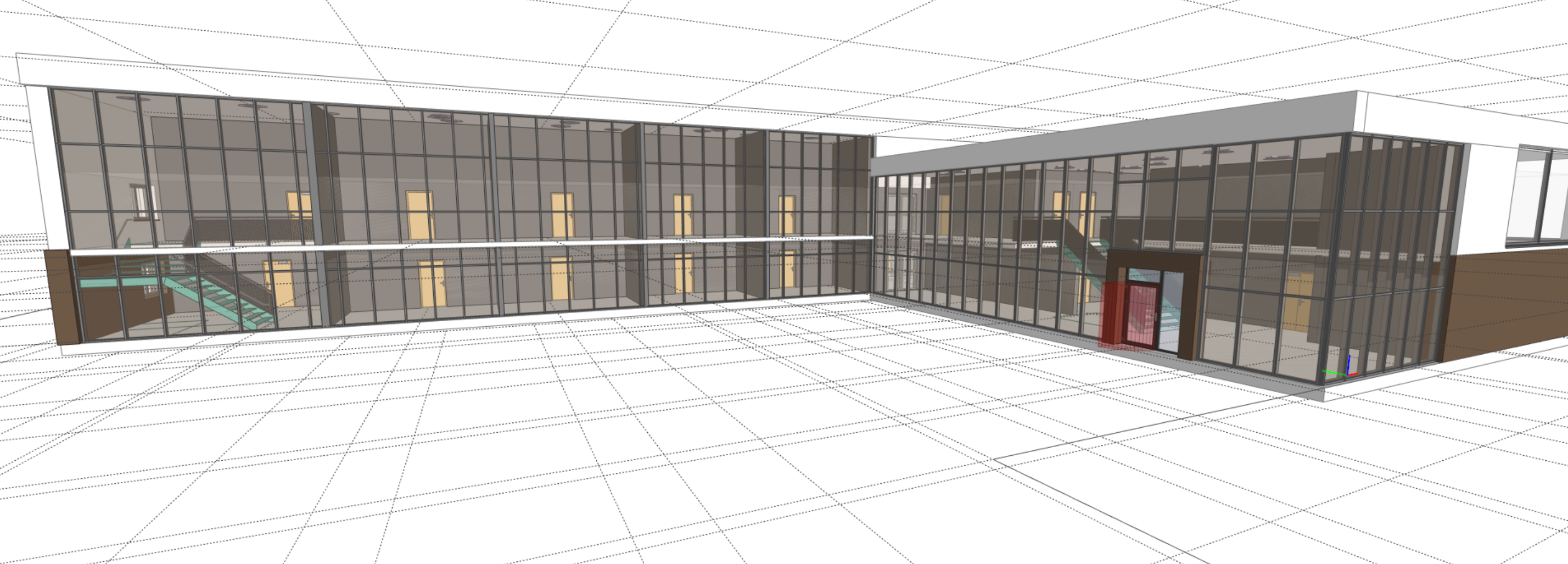}

    \medskip
    \ifccell{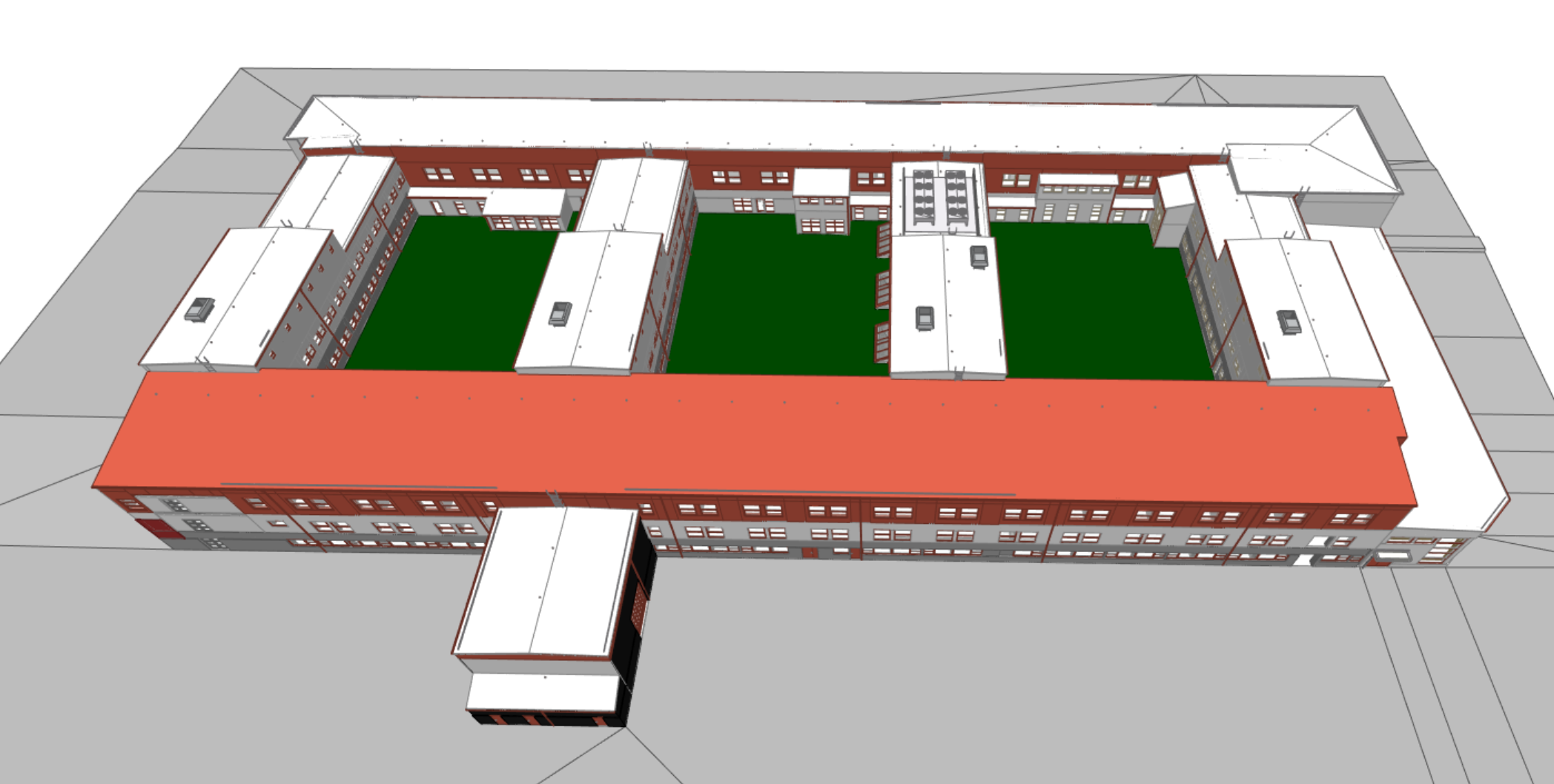}\hfill
    \ifccell{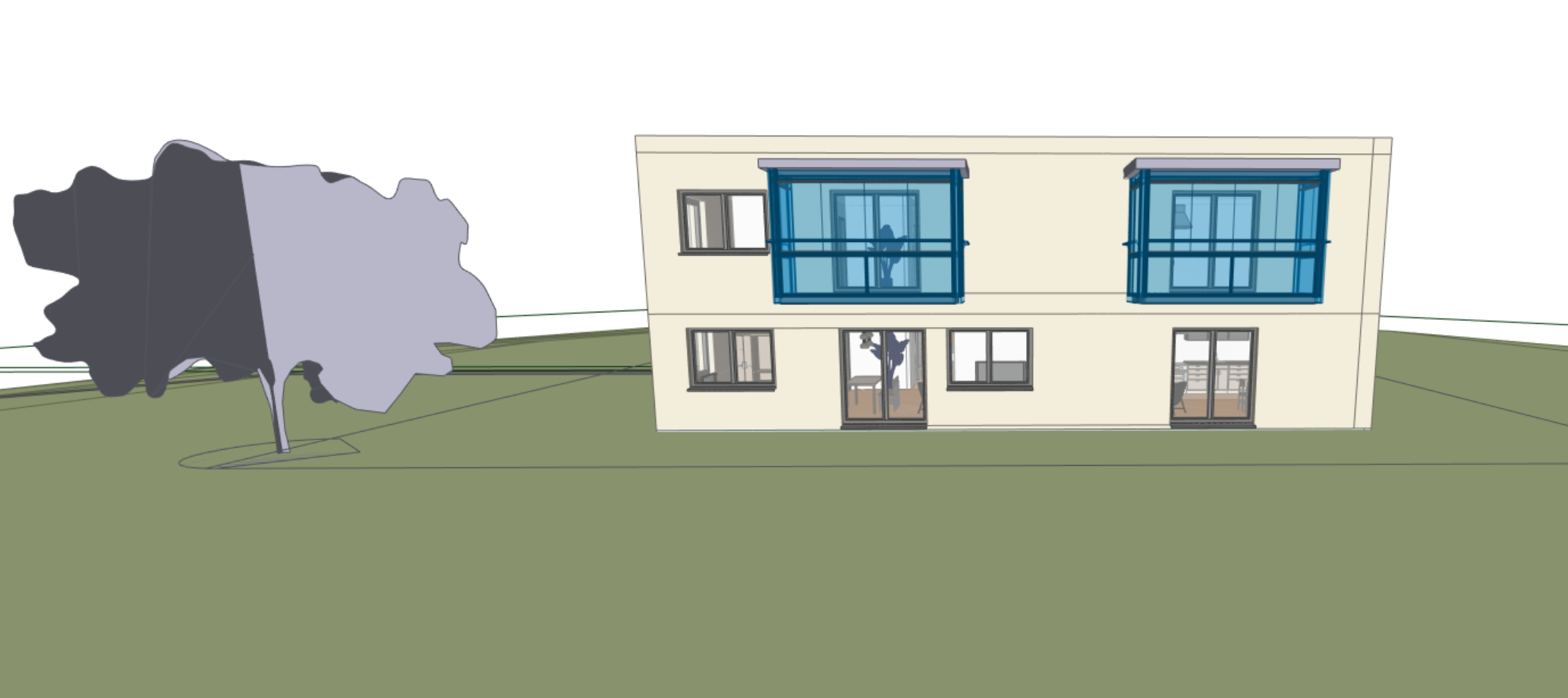}\hfill
    \ifccell{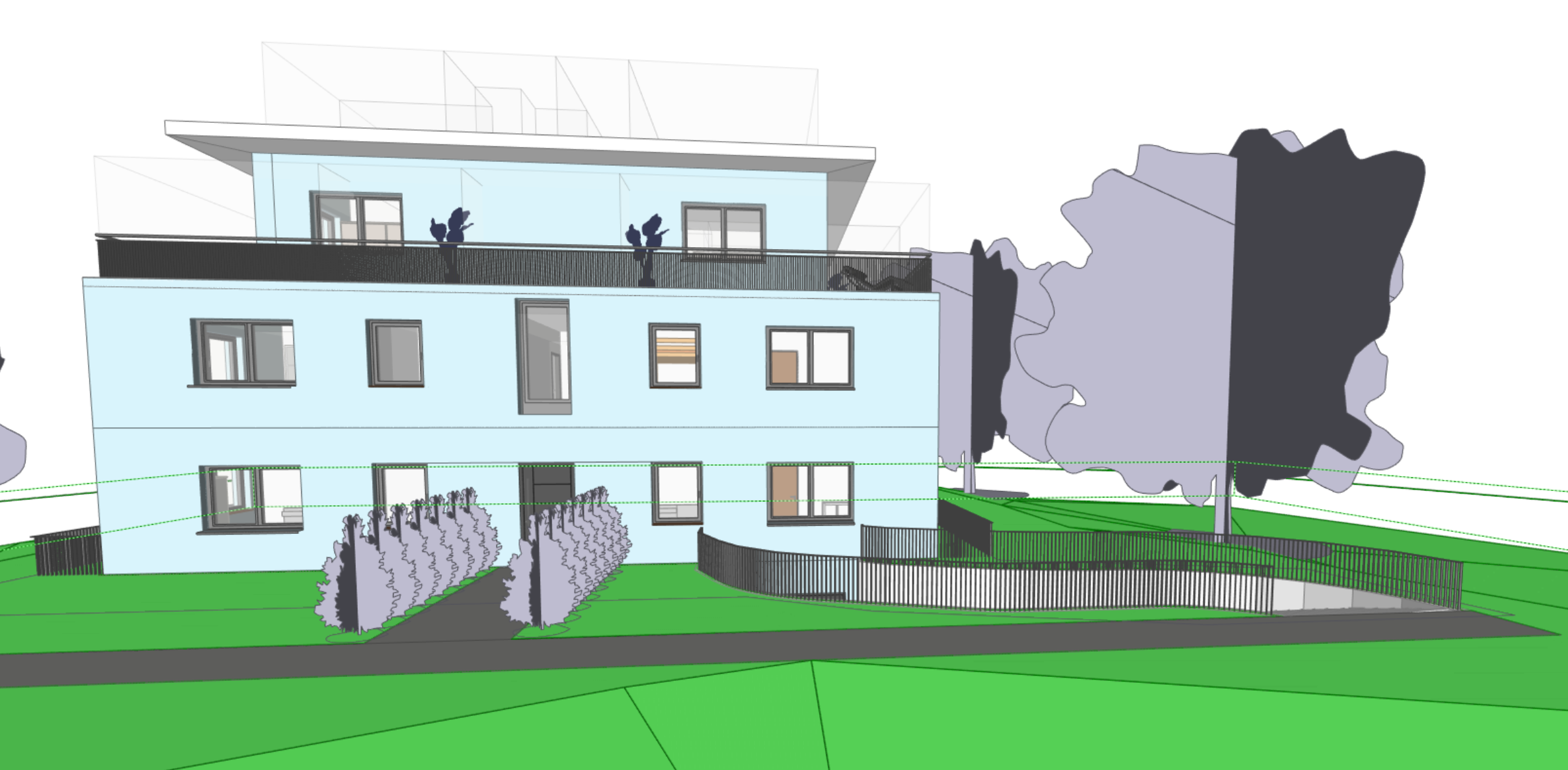}

    \medskip
    \ifccell{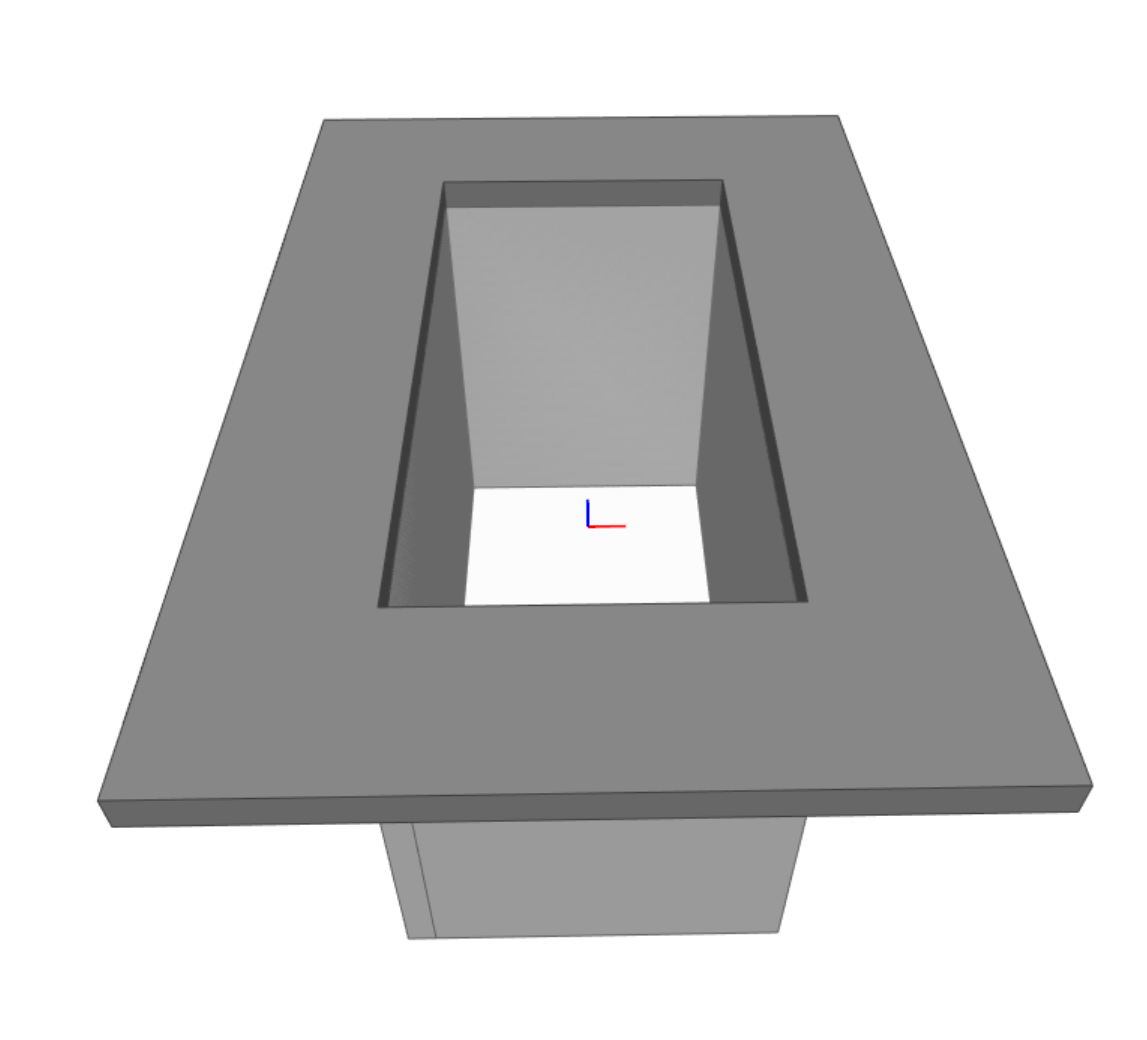}\hfill
    \ifccell{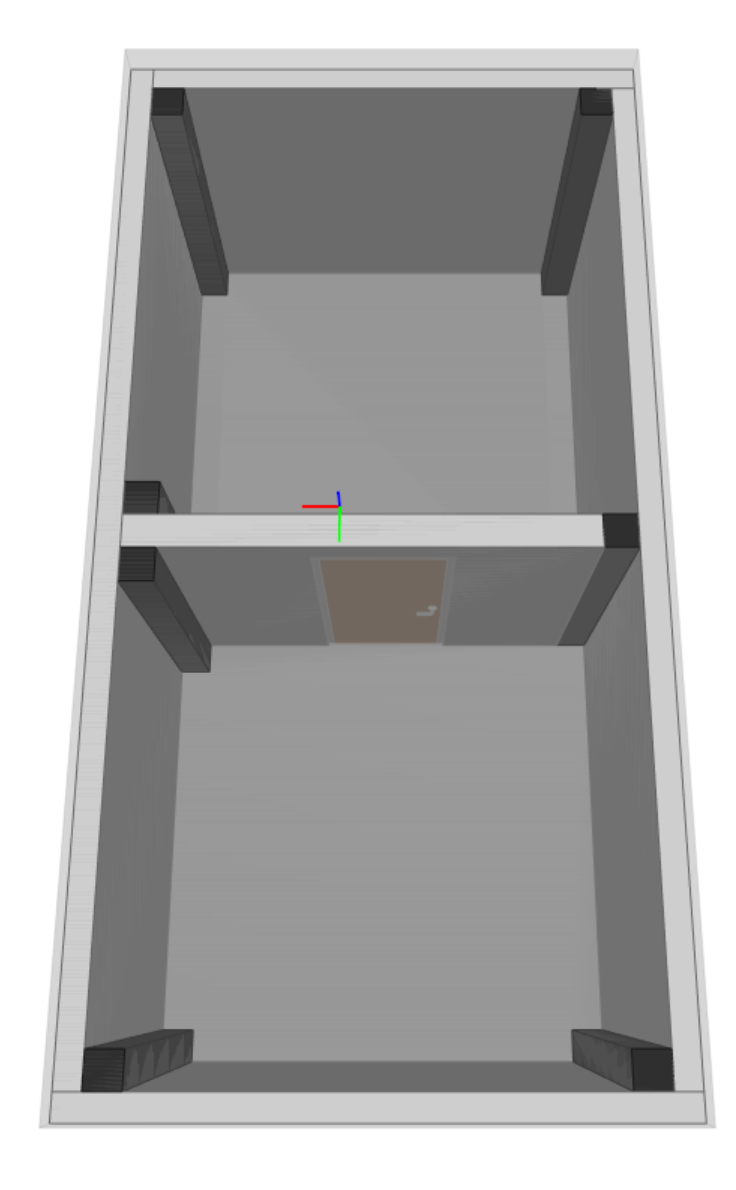}\hfill
    \ifccell{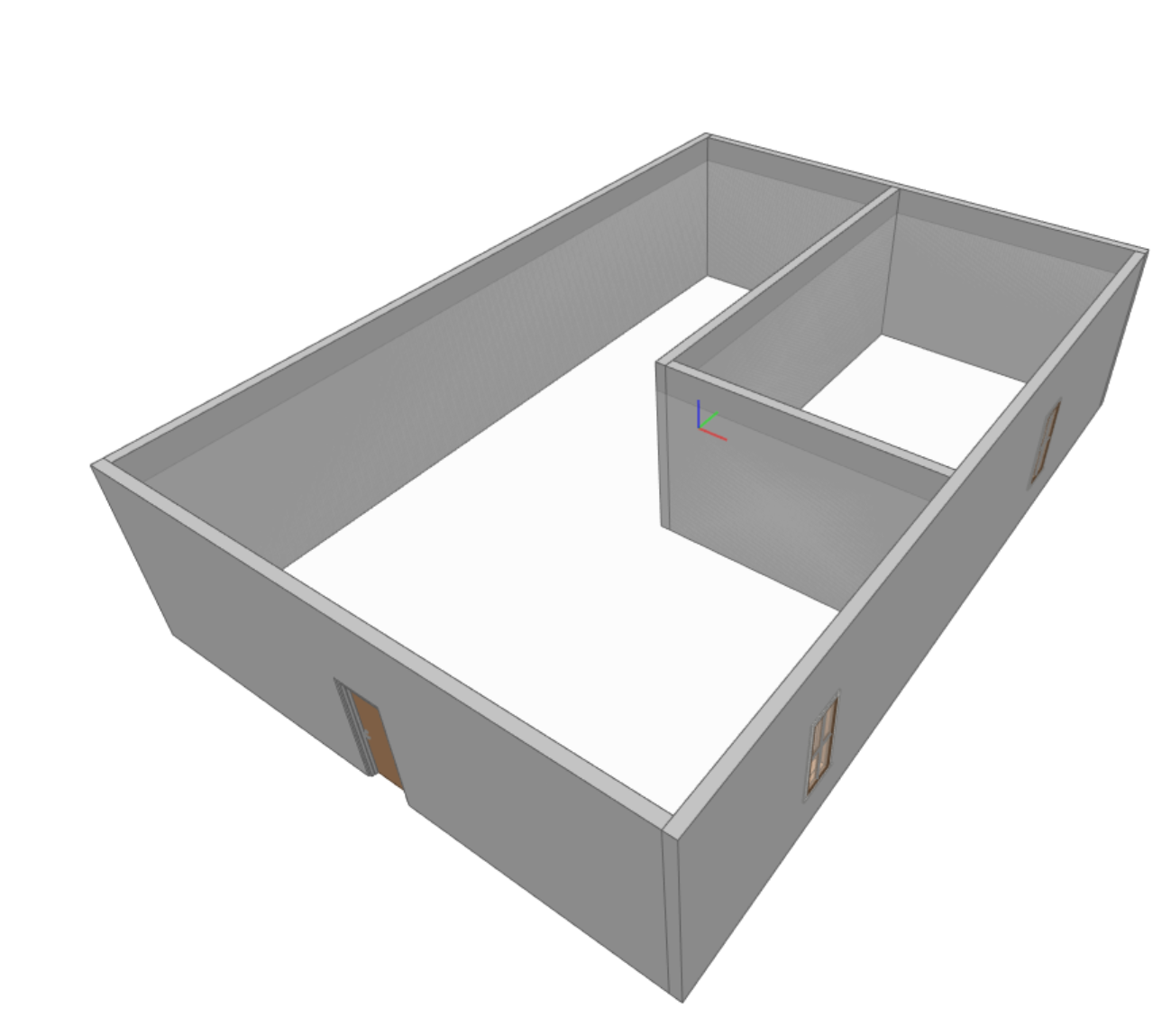}

    \medskip
    \ifccell{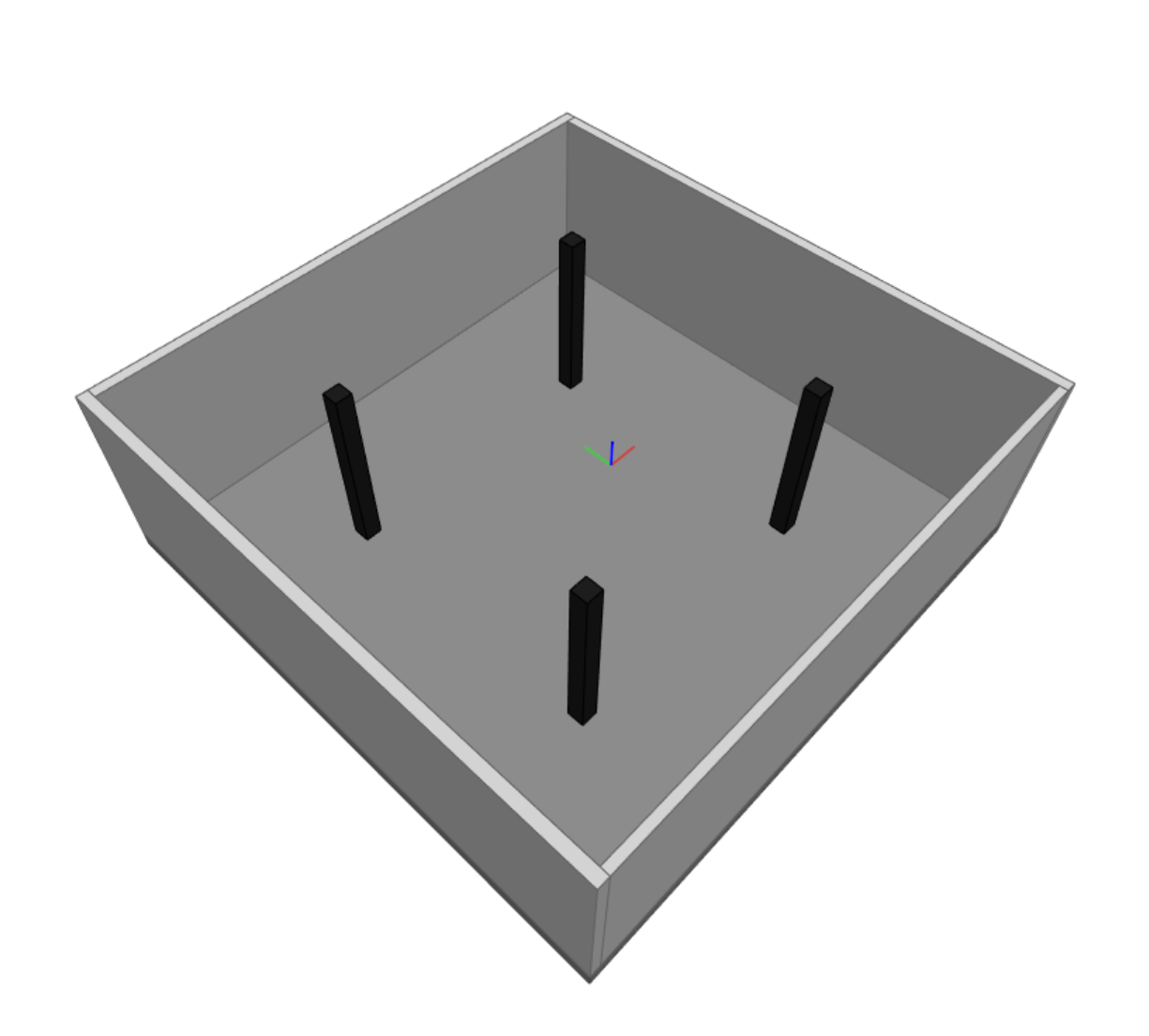}\hfill
    \ifccell{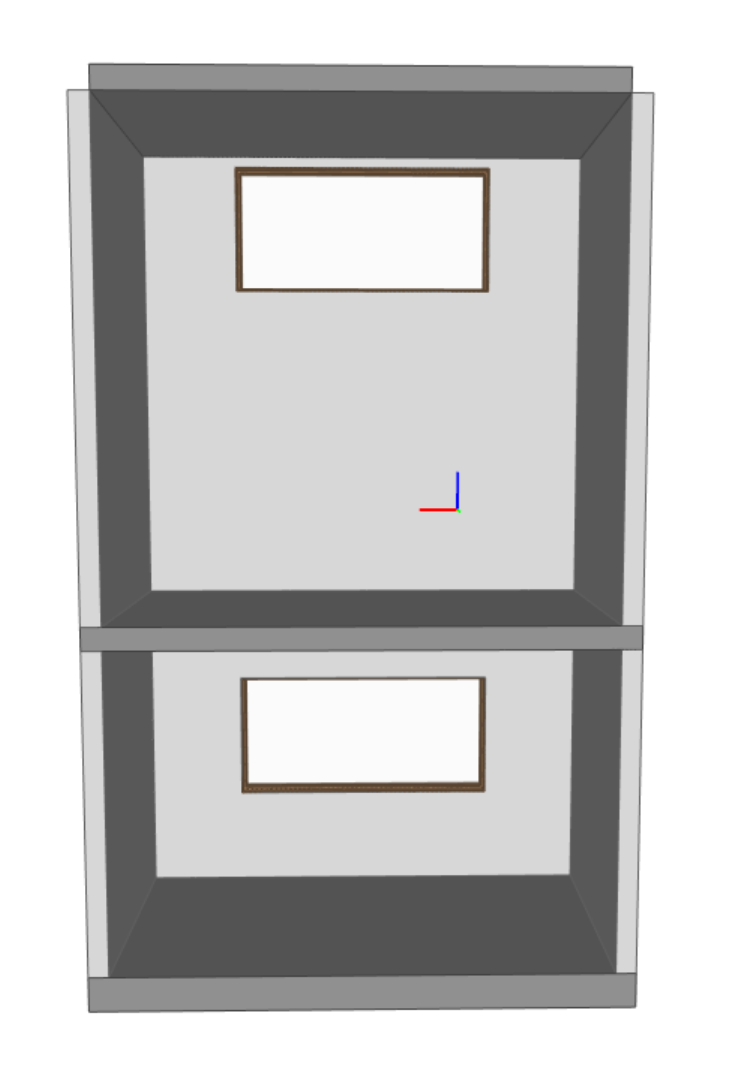}\hfill
    \ifccell{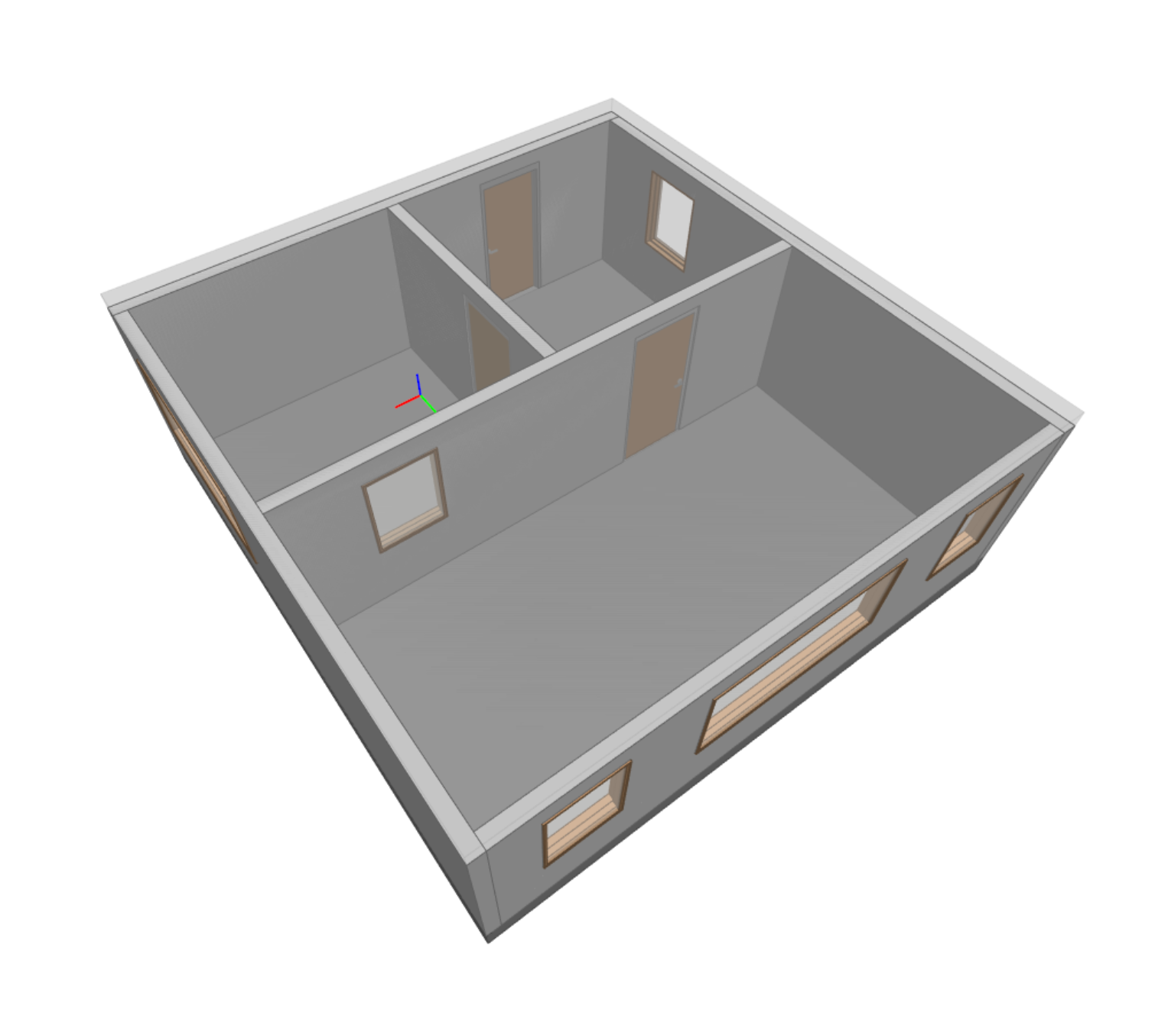}

    \caption{Example IFC files used in the benchmark. The top 6 are realistic large-scale models, and the bottom 6 are smaller synthetic models specifically created for this benchmark.}
    \label{fig:main_4x3_grid}
\end{figure*}

\section{Licenses and Asset Terms}
\label{app:licenses}

The release package includes the license file for the code. The inference harness and the evaluator is released under the MIT License. Benchmark prompts, task metadata, author-created artificial IFC files, and author-created realistic IFC files are released under CC-BY~4.0.

\newpage

\end{document}